\documentclass{article}

\usepackage[final]{neurips_2024}
\setcitestyle{numbers,square,comma}

\usepackage[utf8]{inputenc} %
\usepackage[T1]{fontenc}    %
\usepackage[svgnames]{xcolor}         %
\usepackage[colorlinks=true, citecolor=Navy]{hyperref}       %
\usepackage{url}            %
\usepackage{booktabs}       %
\usepackage{amsfonts}       %
\usepackage{nicefrac}       %
\usepackage{microtype}      %
\usepackage{amsmath}
\usepackage{graphicx}
\usepackage{acronym}
\usepackage{subcaption}
\usepackage{placeins}
\usepackage{color}
\usepackage{fancyvrb}
\usepackage{framed}
\usepackage{listings}

\usepackage{algorithm}
\usepackage{algpseudocode}

\usepackage{wrapfig}
\usepackage{tikz}
\usetikzlibrary{shapes,arrows.meta, positioning, fit, backgrounds, calc,decorations.pathreplacing}

\title{Normalization Layer Per-Example Gradients are Sufficient to Predict Gradient Noise Scale in Transformers}

\author{%
     Gavia Gray \\
     Cerebras Systems \\
     Toronto, Canada \\
     \texttt{gavia.gray@cerebras.net} \\
     \And
     Aman Tiwari \\
     subjective.dev \\
     London, UK \\
     \And
     Shane Bergsma \\
     Cerebras Systems \\
     Toronto, Canada \\
     \And
     Joel Hestness \\
     Cerebras Systems \\
     Sunnyvale, CA \\
}

\newacro{gns}[GNS]{Gradient Noise Scale}
\newacro{ddp}[DDP]{Distributed Data Parallel}
\newacro{cnn}[CNN]{Convolutional Neural Network}
\newacro{sogns}[SOGNS]{Scaled Output Gradient Noise Scale}
\newacro{pepgns}[PEPGNS]{Per-Example Parameter Gradient Noise Scale}
\newacro{tpu}[TPU]{Tensor Processing Unit}
\newacro{tpus}[TPUs]{Tensor Processing Units}
\newacro{ema}[EMA]{Exponential Moving Average}
\newacro{amp}[AMP]{Automatic Mixed Precision}
\newacro{mfu}[MFU]{Model FLOPs Utilization}

\usepackage{shortbold}

\newcommand{\Bcrit}{\mathcal{B}_{\textrm{crit}}}
\newcommand{\Bnoise}{\mathcal{B}_{\textrm{noise}}}
\newcommand{\Bsimple}{\mathcal{B}_{\textrm{simple}}}

\newcommand{\Bbig}{B_{\textrm{big}}}
\newcommand{\Bsmall}{B_{\textrm{small}}}
\newcommand{\Gbig}{G_{B_{\textrm{big}}}} %
\newcommand{\Gsmall}{G_{B_{\textrm{small}}}} %
\newcommand{\Rbb}{\mathbb{R}}
\DeclareMathSymbol{\mlq}{\mathord}{operators}{``}
\DeclareMathSymbol{\mrq}{\mathord}{operators}{`'}
\newcommand{\sqn}[1]{\left\lVert#1\right\rVert_2^2}
\newcommand{\norm}[1]{\left\lVert#1\right\rVert_2}

\begin{document}
\maketitle

\begin{abstract}
    Per-example gradient norms are a vital ingredient for estimating gradient
    noise scale (GNS) with minimal variance. Observing the tensor contractions
    required to compute them, we propose a method with minimal FLOPs in 3D or
    greater tensor regimes by simultaneously computing the norms while computing
    the parameter gradients. Using this method we are able to observe the GNS of
    different layers at higher accuracy than previously possible. We find that
    the total GNS of contemporary transformer models is predicted well by the
    GNS of only the normalization layers. As a result, focusing only on the
    normalization layer, we develop a custom kernel to compute the per-example
    gradient norms while performing the LayerNorm backward pass with zero
    throughput overhead. Tracking GNS on only those layers, we are able to guide
    a practical batch size schedule that reduces training time by 18\% on
    a Chinchilla-optimal language model.

\end{abstract}

\section{Introduction}\label{introduction}

\vspace{-0.45em}

The gradients gathered during the backward pass while training a neural network
are typically inspected via their Frobenius norm, the magnitude of the vector.
This gradient vector may be viewed as the sum of gradients computed over each
individual example in the minibatch. Each of these has its own norm. In this work,
we develop a method to access these norms that works at any scale, for three
common layer types in deep learning models: linear, normalization and embedding
layers.

\begin{figure}
    \centering
    \begin{tikzpicture}[node distance=0.5cm]
  \pgfmathsetseed{21}
  \tikzset{
    bigarrow/.style={
      decorate,
      decoration={brace, amplitude=10pt, raise=2pt},
      thick, -latex, -
    }
  }
  \tikzset{layer/.style={draw, rectangle, minimum height=0.8cm, minimum width=1.2cm, align=center}}
  \tikzset{llabel/.style={
    inner sep=0.0cm
  }}
  \tikzset{bigbox/.style={draw, rectangle, inner sep=0.15cm}}
  \tikzset{worker/.style={bigbox, label=above:Minibatch}}
  \tikzset{aggregator/.style={bigbox, label=above:Aggregated}}
  \tikzset{gradient/.style={->, thick, red}}

  \foreach \w in {0,1,2} {
    \node[layer] (layer1-\w) at (2.3*\w, 0) {};
    \node[llabel, left=of layer1-\w, xshift=0.4cm] (label1-\w) {$l_1$};

    \node[layer, below=of layer1-\w] (layer2-\w) {};
    \node[llabel, left=of layer2-\w, xshift=0.4cm] {$l_2$};
    \node[layer, below=of layer2-\w] (layer3-\w) {};
    \node[llabel, left=of layer3-\w, xshift=0.4cm] {$l_3$};

    \ifnum\w=0
      \def\mylength{0.15cm}
    \else
      \ifnum\w=1
        \def\mylength{0.4cm}
      \else
        \def\mylength{0.28cm}
      \fi
    \fi
    \foreach \x in {1,2,3} {
      \pgfmathsetmacro{\Angle}{random(0,360)}
      \draw[gradient] (layer\x-\w.center) -- ++(\Angle:\mylength);
    }

    \node[worker, fit=(label1-\w) (layer1-\w) (layer3-\w)] (worker-\w) {};
  }

  \draw[bigarrow] ([xshift=0.2cm]worker-2.north east) -- ([xshift=0.2cm]worker-2.south east) {};

  \foreach \w in {3} {
    \node[layer] (layer1-\w) at (2.5*\w, 0) {};
    \node[llabel, left=of layer1-\w, xshift=0.4cm] (label1-\w) {$l_1$};

    \node[layer, below=of layer1-\w] (layer2-\w) {};
    \node[llabel, left=of layer2-\w, xshift=0.4cm] {$l_2$};
    \node[layer, below=of layer2-\w] (layer3-\w) {};
    \node[llabel, left=of layer3-\w, xshift=0.4cm] {$l_3$};

    \draw[gradient] (layer1-\w.center) -- ++(0.11,0.11);
    \draw[gradient] (layer2-\w.center) -- ++(0.08,0.08);
    \draw[gradient] (layer3-\w.center) -- ++(-0.1,-0.1);
    \node[aggregator, fit=(label1-\w)(layer1-\w) (layer3-\w)] (worker-\w) {};
  }
\end{tikzpicture}
    \caption{%
    Gradient noise scale (GNS) is typically computed by comparing per-minibatch
    (aggregated-across-layers) gradients to gradients ``Aggregated'' across
    minibatches.  We estimate GNS with lower variance by making each minibatch
    a single example, and maintain per-layer GNS estimates.  We find the
    magnitude of gradients (visualized by the length of red arrows) to be
    consistent across layers, enabling overall GNS to be computed very cheaply
    using only gradient stats from LayerNorm layers.}
    \label{fig:diagram}  %
\end{figure}
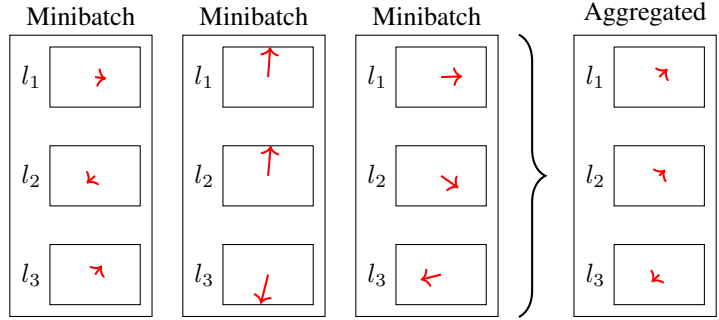

One primary application of a per-example gradient norm is in estimating the
\ac{gns}~\citep{mccandlish2018empirical}, a metric that has been shown to be
useful in training large scale models~\citep{brown2020language}. The uncertainty
of the \ac{gns} estimator depends directly on the size of the batch used to
compute the small batch gradient norm as shown in Section~\ref{gns}. So, the
most precise estimate of the \ac{gns} is obtained by computing the gradient
norms for \emph{each} example in the minibatch: the per-example gradient norm.

To demonstrate \ac{gns} measurement in practice we perform experiments on
contemporary language model architectures, providing a detailed visualisation
of the movement of the \ac{gns} components throughout training, presented
in Section~\ref{experiments}. By inspecting these components it was found
that the \ac{gns} of the model is highly correlated between layer types,
which we give an intuition for in Figure~\ref{fig:diagram}.

However, the practical utility of measuring \ac{gns} with per-example gradient
norms is only present if it can be gathered without affecting training time.
Focusing on LayerNorm~\citep{ba2016layer} layers, we note the main speed
bottleneck is the memory I/O when not implemented as a fused kernel. To
demonstrate this, we develop a custom kernel to compute both the backward pass
and the per-example gradient norms at the same time. Using this kernel the
throughput overhead of gathering the per-example gradient is zero, even
outperforming PyTorch's LayerNorm at larger dimensions. We apply this to
a practical batch size schedule case study in Section~\ref{batch-size}.

To reiterate, the contributions of this work are:

\begin{itemize}
    \item A minimal FLOP algorithm and implementation for computing gradients
        and per-example gradient norms of linear layers
        simultaneously.\footnote{Similar algorithms for other layer types
        described in Appendix~\ref{app-sim}}
    \item Observations that the measured \ac{gns} for LayerNorm layers is highly
        correlated with the GNS of the remaining layers.
    \item Development of an example kernel to implement tracking the \ac{gns} of
        LayerNorm layers that does not affect network throughput (tokens/sec).
    \item Demonstration of a real application of \ac{gns} tracking in a batch
        size schedule experiment that obtains an 18\% wall-time speedup in
        training a Chinchilla-optimal~\citep{hoffmann2022training} LLM.
\end{itemize}

\section{Background}\label{background}

\subsection{Gradient Noise Scale}\label{gns}

\ac{gns} is a metric derived from observing a second order Taylor expansion of
the change in a loss function under the following assumption on the noise in the
gradient estimate~\citep{mccandlish2018empirical},
\begin{align}
    G_{\text{est}(\theta)} \sim \mathcal{N} \left(G (\theta), \frac{1}{B} \Sigma (\theta) \right),
\end{align}
where $G_{\text{est}}$ is the observed gradient, $B$ is the batch size, and $\theta$ the parameters
of the model. Here, $G$ is the unobserved ``true'' gradient and $\Sigma$ is the
covariance of the gradient estimate. The Taylor expansion mentioned is,
\begin{align}
    \mathbb{E}[L(\theta - \epsilon G_{est})] = L(\theta) - \epsilon |G|^2 + \frac{1}{2} \epsilon^2 \left( G^T HG + \frac{tr(H\Sigma)}{B} \right).
\end{align}
Where $\epsilon$ is the learning rate and $H$ is the Hessian of the loss.
On the right hand side is a factor that depends on $B$.
It may be shown~\citep{mccandlish2018empirical} that the optimal step size and optimal change
in the loss is achieved when $B = \Bnoise := tr(H \Sigma) / G^T H G$.
Averaging this optimal step over an entire run,
and measuring this value by a grid search, yields $\Bcrit$ which describes
a batch size that meets an optimal tradeoff between cost and training speed.
It is shown by analysis and experiment that $\Bnoise \approx \Bcrit$.

As this depends on the Hessian, which is typically unavailable,
\citet{mccandlish2018empirical} suggest making the assumption that the Hessian
is diagonal, which yields
\begin{align}
    \Bsimple = \frac{tr(\Sigma)}{G^T G}.
\end{align}

To compute $\Bsimple$ \citet{mccandlish2018empirical} define the
unbiased estimators $\mathcal{S}$ and $\sqn{\mathcal{G}}$ as:
\begin{align}
    \sqn{\mathcal{G}} := \frac{1}{\Bbig - \Bsmall} \left( \Bbig \sqn{ \Gbig } - \Bsmall \sqn{\Gsmall} \right) \approx G^T G \label{eq:g-est}\\
    \mathcal{S} := \frac{1}{1 / \Bsmall - 1 / \Bbig} \left( \sqn{\Gsmall} - \sqn{\Gbig} \right) \approx tr(\Sigma),
    \label{eq:s-est}
\end{align}
where $\Bbig$ and $\Bsmall$ are the batch sizes used to compute the gradients
$\Gbig$ and $\Gsmall$, respectively (potentially corresponding to \emph{Aggregated} and \emph{Minibatch} gradients as depicted in Figure~\ref{fig:diagram}).

$\norm{\Gbig}$ is trivially computed using the gradients accumulated for the
optimizer but $\norm{\Gsmall}$ is not. One option is to use the gradients
communicated between \ac{ddp} nodes, but this has two downsides: (1)~the variance of
the estimate is tied to the \ac{ddp} configuration and (2)~the estimate is not
available in all training configurations. For example, experiments on a single
GPU cannot use this method. One can also access the gradients during gradient
accumulation, but this similarly depends on the training configuration.
A full taxonomy of the options for computing $\norm{\Gsmall}$ is provided in
Appendix~\ref{taxonomy}.

\begin{figure}
    \centering
    \begin{subfigure}[b]{0.39\textwidth}
        \centering
        \includegraphics[width=\textwidth]{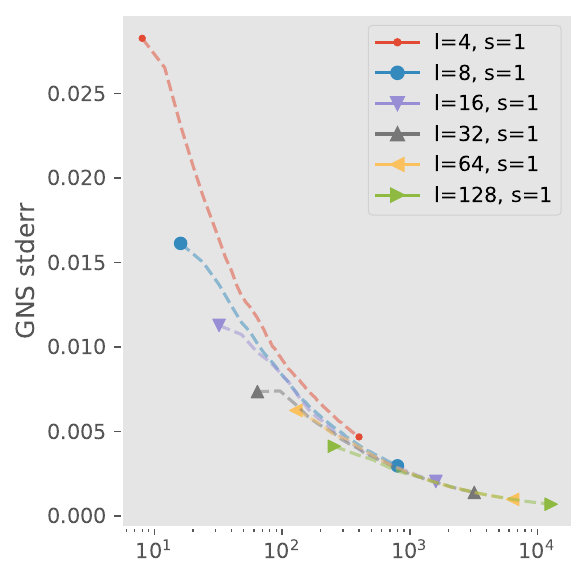}
        \label{fig:bigvariance}
    \end{subfigure}
    \begin{subfigure}[b]{0.39\textwidth}
        \centering
        \includegraphics[width=\textwidth]{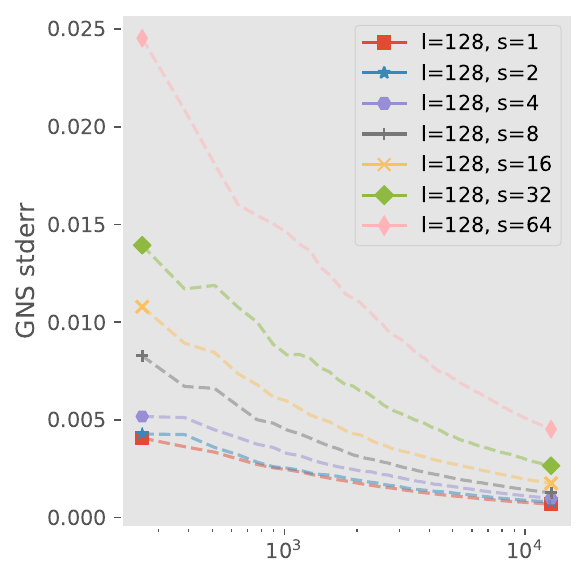}
        \label{fig:smallvariance}
    \end{subfigure}
    \caption{%
        The variance of the \ac{gns} estimator for different $\Bbig$ (left) and
        $\Bsmall$ (right) sizes. $\Bbig = l$ and $\Bsmall = s$ in legends.
        Stderr is estimated using a jackknife resampling method for ratio
        estimators~\citep{choquet1999bootstrap}. For the same number of samples
        processed, a smaller $\Bsmall$ always has a lower standard error, while
        the size of the large batch, $\Bbig$ does not affect the standard error.}
    \label{fig:variance}
\end{figure}

For each observation of $\norm{\Gbig}$ we may observe multiple $\norm{\Gsmall}$,
typically $\Bbig / \Bsmall$ of them. On each step the estimate of
$\sqn{\Gsmall}$ is therefore a mean over $\Bbig / \Bsmall$ samples, whose
variance is reduced according to the law of large numbers. However, the GNS is
a ratio of the unbiased estimators in Equations \ref{eq:g-est} and
\ref{eq:s-est}, so it may not be clear how this affects uncertainty in the \ac{gns}
estimate. Figure~\ref{fig:variance} explores this
relationship by simulation of a setting where the \ac{gns} is set to 1 while
varying $\Bbig$ and $\Bsmall$. We find it is always better (less
uncertainty) to use the smallest possible $\Bsmall$ to estimate the \ac{gns},
while the choice of $\Bbig$ is irrelevant.

\subsection{Efficient Per-example Gradient Norms}\label{per-example}

\citet{goodfellow2015efficient} proposes a trick to compute gradient norms for
individual examples in a minibatch, which would provide the minimum variance
estimate of the \ac{gns} as described in Section~\ref{gns}. Neglecting the
original derivation, by writing
the desired squared norm as a tensor contraction the trick may be reproduced
automatically via \texttt{einsum} path optimization~\citep{smith2018opt_einsum,dangel2020backpack}.
The tensor contraction for per-example gradient norms, $n_{b}^2$, of a linear layer in the
2D setting is,
\[
    n_{b}^2 = \sum_{i,k} (w')^2_{bik} = \sum_{i, k} x_{bi} x_{bi} y'_{bk} y'_{bk},
\]
where $x$ are the activations prior to a linear layer, $y'$ are the gradients
of the loss with respect to the outputs of the linear layer and $w'$ are the
gradients of the loss with respect to the weights of the linear layer.

\citet{li2022large} extend this trick to the three dimensional case.
For inputs $\XB \in \Rbb^{B \times T \times I}$ and outputs $\YB \in \Rbb^{B
\times T \times K}$, the per-example gradient norm $n_{b}$ is,
\[
    n_{b}^2 = (w')^2_{bik} = (\sum_t x_{bti} y'_{btk})^2 = x_{bti} y'_{btk} x_{bui} y'_{buk} = \langle \XB \XB^T, \YB' \YB'^T \rangle_F^2,
\]
which has $O(T^2)$ memory complexity in the sequence length $T$.\footnote{This specific Einstein contraction is not used by \citet{li2022large}
but appears in the Backpack library~\citep{dangel2020backpack} We provide the
vector algebra contraction path chosen by \citet{li2022large} on the right.}.
Index sets are $b \in [1,B], \; i \in [1,I], \; k \in [1,K], \; t,u \in [1,T]$.
At some point, the I/O cost of computing the per-example gradient norms
by computing the full $w'_b$ explicitly will be cheaper.
Noting this fact motivated the work in Section~\ref{sim-per-example} and the
practical relationship between these resource costs is explored in Section~\ref{flops}.

\subsection{Related Work}\label{related-work}

\paragraph{Gradient norms}

One common motivation for computing per-example gradient norms is for
differential privacy.  By bounding the gradient for any single
example, we can ensure each example has a limited impact on the
final parameters~\citep{rochette2019efficient,li2022large}.
Per-example gradient clipping has been performed with convolutional
networks~\citep{rochette2019efficient} and sequential models, e.g.,
LLMs~\citep{li2022large}.
These methods allow control over per-example gradient norms even when
training with large batch sizes.  Approaches like these are
implemented in the differential-privacy library Opacus~\citep{opacus},
and have support natively in PyTorch, but are less efficient than the
methods proposed in this paper.
An alternative mechanism to manifest per-example gradient norms is to
simply use a batch size of one.  While not efficient enough for
training large-scale networks, such sequential training may arise in
situations such as reinforcement learning, where per-example gradient
clipping has also been performed (to improve
stability~\citep{wang2016dueling}).

\paragraph{Gradient noise scale}

The \acl{gns}~\citep{mccandlish2018empirical} has been widely used for
training large-scale neural networks.  For example,
\citet{brown2020language} note the \ac{gns} was measured during
training and used to guide batch sizing when training GPT-3\@.
\citet{dey2023cerebrasgpt} mention that operating near the critical
batch size, as dictated by the \ac{gns}, is important for
hyperparameter transfer under the maximal update
parameterization~\cite{yang2022mup}.
Even when not explicitly mentioned in publications, open source code
often implements the \ac{gns} (e.g., see
codebases~\cite{eleuther2024gptneox,crowson2024kdiffusion} for
GPT-NeoX~\cite{black2022gptneox} and Hourglass Diffusion
Transformer~\cite{crowson2024scalable}).

Measurements similar to the \ac{gns} have also been used in a range of
prior work to guide batch sizing for minibatch
SGD~\citep{byrd2012sample,de2016big,balles2016coupling,yin2018gradient}.
\citet{chen2018effect} show experimentally that wider networks can be
trained using larger batches; they also establish a theoretical
connection between wider networks and gradient variance, albeit for
simple two-layer networks.  In contrast, \citet{shallue2019measuring}
found empirically that \emph{narrower} Transformers scale better to
larger batch sizes.
\citet{smith2017bayesian} propose a noise scale based not on gradient
variance, but on the learning rate, dataset size, and batch size (similar to the notion of temperature in Section~\ref{temperature}).
\citet{zhang2019algorithmic} find the critical batch size depends on
the choice of optimizer.
\citet{faghri2020study} introduce a gradient clustering and stratified
sampling approach to minimize minibatch gradient variance, and use
this approach as a tool to help understand optimization.

\paragraph{Gradient variance}

Beyond computing the \ac{gns}, our method can support other
applications where measuring the distribution of per-example gradients
is useful or informative.
Gradient variance has been used to classify the \emph{difficulty} of
examples~\citep{agarwal2022estimating}, which can be used, for
example, to surface problematic examples for human auditing.
The question of whether gradient distributions tend toward Gaussian in
the (central) limit is of theoretical
significance~\citep{smith2017bayesian}, with implications toward the
ability of SGD to escape sharp minima and land in wide
basins~\citep{zhu2018anisotropic,nguyen2019first,simsekli2019tail}.
Bounded gradient variance is also assumed in some convergence
analysis~\citep{bottou2018optimization,zhang2022adam}, as noted
in~\citep{faghri2020study}.

Perhaps the most familiar use of gradient variance is of course in
adaptive optimizers like Adagrad, Adam, and others that reduce step
sizes in high-gradient-noise
directions~\citep{duchi2011adaptive,zeiler2012adadelta,schaul2013no,kingma2014adam,reddi2019convergence}.
\citet[App.~C]{hilton2022batch} directly relate Adam second moment statistics
to a \emph{component-wise} version of the \ac{gns}.
Optimizers typically estimate gradients jointly across training steps
and minibatches, however vSGD~\citep{schaul2013no} leverages separate
components for gradient momentum and for gradient variation across
samples.
\citet{zhang2020adaptive} find the variance of gradient norms across
examples predictive of whether vanilla SGD outperforms adaptive
optimizers, however recent work has shown
Adam to outperform SGD even in the (noise-free) full gradient descent
setting~\cite{kunstner2023noise,kunstner2024heavy}.

\section{Simultaneous Per-example Gradient Norms}\label{sim-per-example}

As described in Section~\ref{background}, computing \ac{gns} requires small
batch gradient norms. Typically, these may be gathered during gradient
accumulation or \ac{ddp} communication.\footnote{A complete taxonomy for small batch gradient computation is given in Appendix~\ref{taxonomy}.}
However, these methods are not universally applicable and may not be available
in all training configurations. In this section we describe a method for baking
the computation of the per-example gradient norms into the computation graph,
making it universally applicable.
The typical tensor contraction used to compute the backward gradient in a linear layer using
the input activations, $\xB$, and gradients, $\gB$, is,
\[
    w_{k, l}' = \sum x_{\ldots k} g_{\ldots l},
\]
in other words, a sum over vector outer products for every vector in the
trailing dimension. In principle, it is possible to
access the intermediate tensor containing the batch dimension $w_{b k l}' = \sum
x_{b \ldots k} g_{b \ldots l}$. This allows us to compute the per-example gradient
norms with FLOPs scaling at the same rate as the normal, non-per-example backward pass (Figure~\ref{fig:gns-flops}), albeit
at increased I/O cost due to having to materialize the intermediate tensor.

A generic algorithm to compute the per-example gradient norms simultaneously
with the weight gradient in a standard linear layer is provided in
Algorithm~\ref{alg:linear-layer} using \texttt{einsum} for readability and
portability.\footnote{Additional algorithms for Embedding and
LayerNorm layers are described in Appendix~\ref{app-sim}.} The reason for the
correction in step 4 can be seen by considering the gradient of loss function
$L$ with respect to the weights on a single example $b$, $w_b$,
\[
    \nabla_{w_b} \frac{1}{B} \sum_b L(x_b) = \frac{1}{B} \nabla_{w_b} L(x_b),
\]
computing the squared norm of this will therefore contain a factor of $1/B^2$,
which must be corrected for.

\begin{algorithm}
\caption{Linear Layer Simultaneous Per-Example Gradient Norm Computation}
\label{alg:linear-layer}
\begin{algorithmic}[1]
\Require gradient tensor $\gB$ of shape $(B, ..., L)$, input activation tensor $\xB$ of shape $(B, ..., K)$
\Ensure weight gradient tensor $\wB'$ of shape $(K, L)$, mean of per-example squared norms $\sqn{\wB_b'}$
\State $\wB_b' \gets \text{einsum}(\mlq b...k,b...l \rightarrow bkl \mrq, \xB, \gB)$
\State $\mathbf{s}_{w} \gets \text{einsum}(\mlq b k l \rightarrow b \mrq, \wB_b'^2)$
\State $\wB' \gets \text{einsum}(\mlq b k l \rightarrow k l \mrq, \wB_b')$
\State $\sqn{\wB_b'} \gets 1/B \times \text{einsum}(\mathbf{s}_w, \mlq b \rightarrow \mrq) \times B^2$ \# reduce by mean then apply correction
\State \Return $\wB', \sqn{\wB_b'}$
\end{algorithmic}
\end{algorithm}

\subsection{FLOPs and I/O Costs}\label{flops}

\begin{figure}
    \centering
    \includegraphics[width=0.45\linewidth]{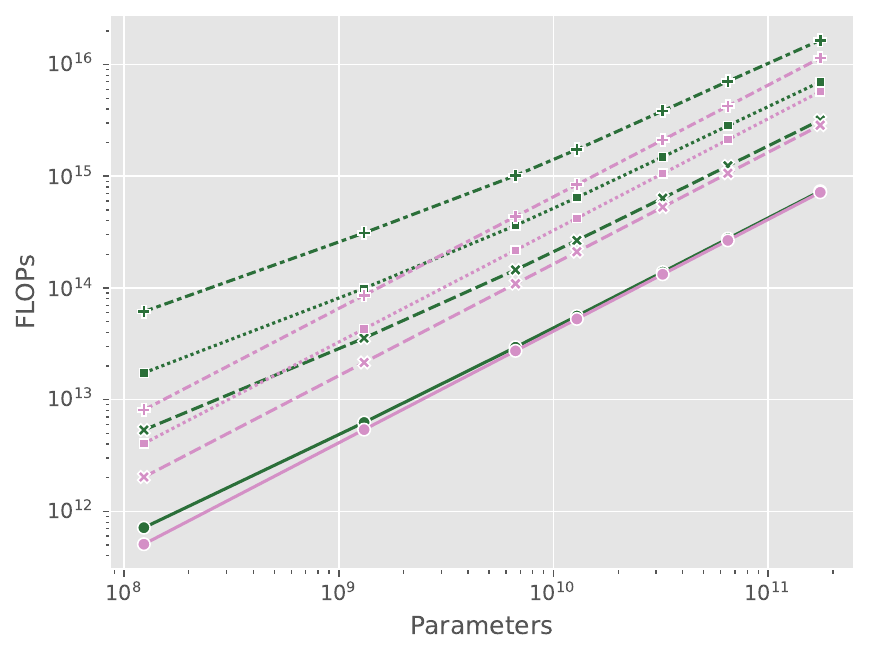}
    \quad
    \includegraphics[width=0.45\textwidth]{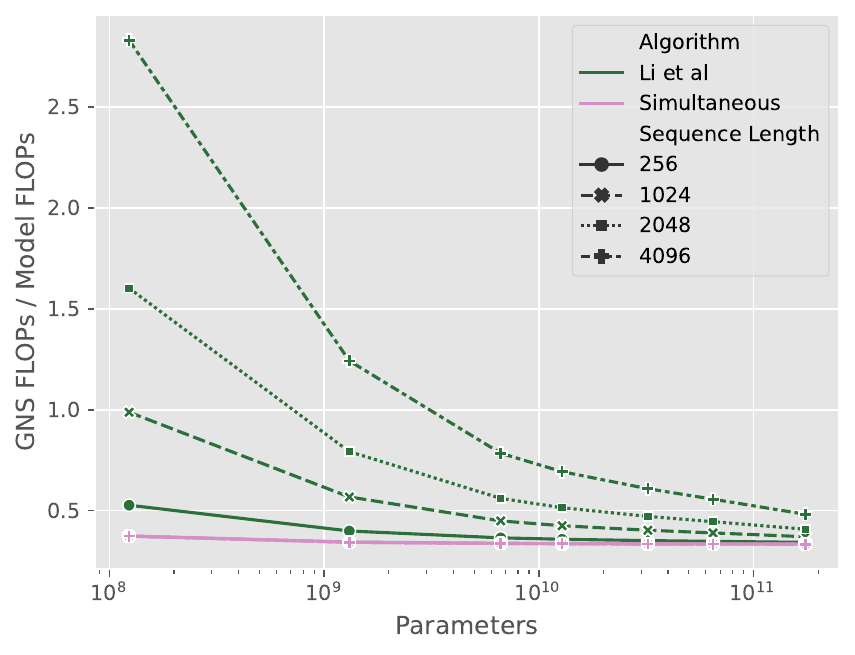}
    \caption{%
        FLOP cost of computing per-example gradient norms. (Left) Total FLOP
        cost. (Right) Proportional cost versus one model forward and backward
        pass. The FLOP cost of Simultaneous per-example gradient norms is
        strictly dominant to alternative methods (left) and the ratio of this
        additional cost to the FLOP cost of processing the entire model does not
        depend on context length (right).}
    \label{fig:gns-flops}
\end{figure}

\begin{figure}
    \centering
    \includegraphics[width=0.6\linewidth]{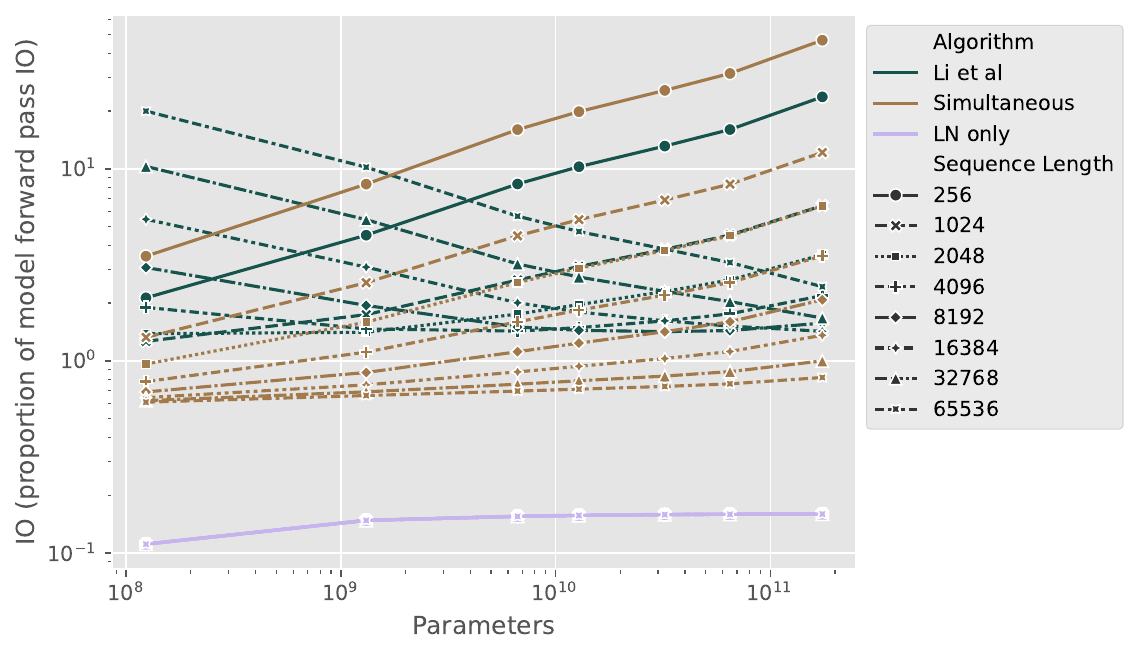}
    \caption{%
        Total I/O cost of computing per-example gradient norms, assuming
        gradients and parameters are stored with 4 bytes of precision.
        The relative IO cost of Simultaneous per-example gradient norms is
        less than \citet{li2022large} for very long contexts for all model
        scales, approximately equivalent for models of 10B parameters and
        4096 context length, and higher for shorter contexts with larger
        models. The IO cost of LN (LayerNorm) per-example gradient norms alone
        is much lower than either method.}
    \label{fig:total-gns-io}
\end{figure}

The computational cost of computing per-example gradient norms can be broken
down into FLOPs, in Figure~\ref{fig:gns-flops}, and I/O, in
Figure~\ref{fig:total-gns-io}, with matrix multiplication on current devices
being potentially bottlenecked by both. We estimate ideal FLOP and DRAM I/O
costs, assuming optimal reuse of data loaded from DRAM into SRAM with no
recomputation. In practice, duplicate computation may be used to improve
wall-clock time and to fit within hardware limitations of the amount of shared
memory available. We compare here against the efficient per-example gradient
norm method described by \citet{li2022large}, which the authors note is only
efficient (in terms of I/O cost) when $2 T^2 < P D$, where $T$ is the sequence
length, $P$ is input and $D$ is output dimension of the linear layer. This
bound is discussed further in Appendix~\ref{app-flops}.

In terms of FLOPS, Figure~\ref{fig:gns-flops} shows the simultaneous
per-example gradient norms are almost always preferable, only being more expensive
for very short sequence lengths in small models. The reason for this is shown
on the right hand side; the number of FLOPs required to compute the simultaneous
per-example gradient norms is independent of the sequence length.

The I/O cost shown in \ref{fig:total-gns-io} illustrates a tradeoff in computing
the per-example gradient norm. The simultaneous method is more expensive at
large model sizes with short sequence length because it must act on a large intermediate tensor.

To estimate model flops, we use PyTorch's
FLOPCounterMode, which only measures the FLOPs in matrix multiplications and
attention computation, however these make up the vast majority of the FLOPs in
a Transformer model.

\section{Gradient Noise Scale in Transformer Language Models}\label{experiments}

Using the methods described in previous sections to measure per-example gradient
norms and estimate the \ac{gns}, we perform experiments on a 111M parameter
Chinchilla-optimal language model~\citep{dey2023cerebrasgpt,hoffmann2022training} using the
OpenWebText dataset~\citep{gokaslan2019owt}.\footnote{The code to replicate these experiments may be found at \url{https://github.com/CerebrasResearch/nanoGNS/tree/main/exact}.}
As the prior work was performed on
Pile~\citep{gao2020pile}, Appendix~\ref{optimal} describes an experiment to
check the optimality of the Chinchilla model on this dataset. We also found
Flash attention led to numerical instability, which we were able to mitigate with
an architectural modification described in Appendix~\ref{instability}.

All experiments computed
per-example gradient norms for all layers in the model with the exception of the
performance results of Sections~\ref{universal} and~\ref{bss}, which only
computed per-example gradient norms for the normalization layers. Each
experiment was run on Nvidia A10 GPUs, in either 12 or 24 hours depending on the
precision used, Bfloat16 or Float32 respectively. We used the
nanoGPT\footnote{\url{https://github.com/karpathy/nanoGPT}} codebase with the
layers described in Section~\ref{sim-per-example} added.

\begin{figure}
    \centering
    \includegraphics[width=0.8\textwidth]{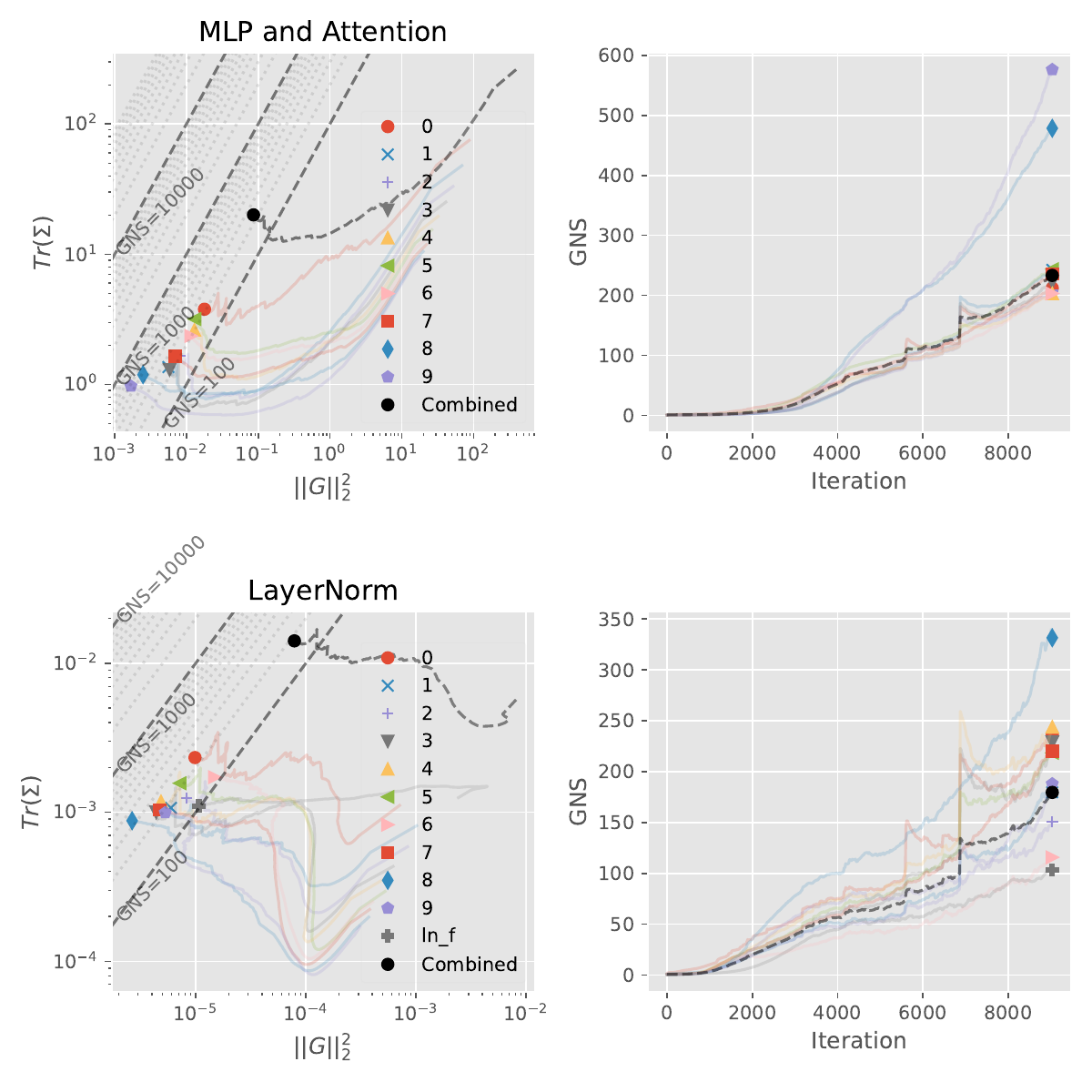}
    \caption{%
        GNS phase plot: Linear/Embedding layers are separated from LayerNorm
        layers by row. Component estimators of Equations~\ref{eq:g-est}
        and~\ref{eq:s-est} are shown (left) with the GNS over the course of
        training on the (right).}
    \label{fig:gns_by_index}
\end{figure}

Having an accurate estimate of the \ac{gns} statistics $\sqn{\mathcal{G}}$ and
$\mathcal{S}$ allows us to visualize the movement of both in a phase space
during training as shown in Figure~\ref{fig:gns_by_index}. LayerNorm layers
are separate from the rest of the network because their statistics are much
smaller and to illustrate how the resulting GNS estimates on the right track
each other. To observe these trends in another training regime, see
Figure~\ref{fig:gns_by_index_bss} in Appendix~\ref{app-gns}.

\subsection{The Temperature of Training}\label{temperature}

\begin{figure}
    \centering
    \includegraphics[width=0.6\textwidth]{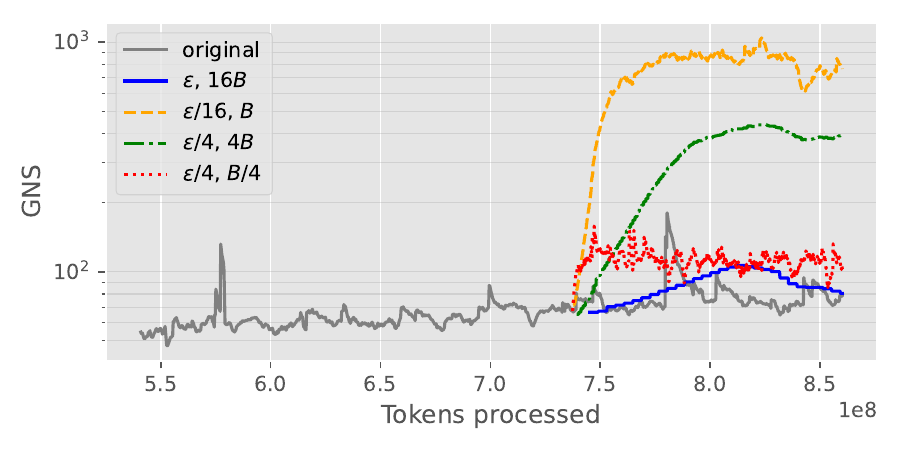}
    \caption{
        During the middle of training a 111M parameter language model on
        OpenWebText, the learning rate, $\epsilon$ or batch size, $B$ were
        varied, restarting the run from the same point. This Figure replicates
        an experiment from \citet{mccandlish2018empirical} showing how varying the ratio
        causes changes in the measured GNS, but here only due to changes in the
        learning rate. Changes in the batch size do not have the predicted effect.}
    \label{fig:temp}
\end{figure}

\citet[App. C]{mccandlish2018empirical} observed that the \ac{gns} measurement
depends on the batch size and learning rate used in training. In fact, from the
derivation outlined in Section~\ref{gns}, the gradient noise scale is only
well-defined at the optimal learning rate. Using a toy model of a quadratic
loss function, they observed that the \ac{gns} should be inversely proportional
to the temperature, $T$, a ratio of batch size $B$ to learning rate $\epsilon$:
\[
    \Bnoise \propto \Bsimple \propto \frac{1}{T} = \frac{B}{\epsilon}.
\]
This enables a testable prediction that the \ac{gns} will increase with increasing
batch size or with descending learning rate. This prediction was found to accurately
describe experiments on a small convolutional model on the SVHN dataset. We repeat it here
in the setting described above in Figure~\ref{fig:temp}. To match the results of
\citet{mccandlish2018empirical}, all interventions tested should yield the same
result. We find the \ac{gns} does indeed react predictably to changes in
the learning rate, but the reactions to changes in the batch size are not
predicted by the theory.

\subsection{GNS Correlates Between Layer Types}\label{correlation}

Inspection of Figure~\ref{fig:gns_by_index} suggests the LayerNorm layers
produce a similar GNS, when combined, as the total GNS of the model. Before
describing how to quantify this relationship we must first note that the
unbiased estimators $\sqn{\mathcal{G}}$ and $\mathcal{S}$ are noisy. All \ac{gns}
figures presented in this paper and other work smooth both of these estimators,
typically with an \ac{ema} filter, before computing the \ac{gns}
ratio.\footnote{The results described in Figure~\ref{fig:gns_by_index} are
    explored at a 1.3B parameter scale in Appendix~\ref{app-larger}.}

\begin{figure}
    \centering
    \includegraphics[width=0.8\textwidth]{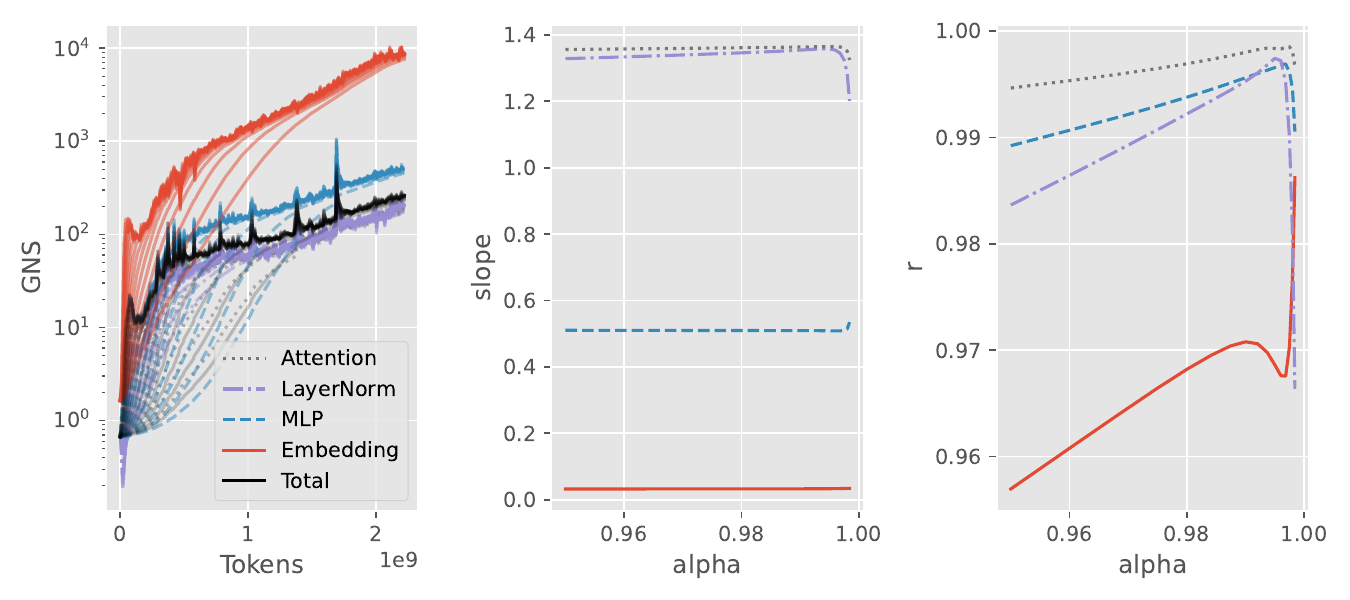}
    \caption{%
        Regression of total GNS using the GNS of each layer type. (Left) GNS of
        each layer type and the total GNS are plotted against the number of
        tokens processed for varying \ac{ema} alpha settings. (Center \& Right)
        The slope and Pearson's correlation coefficient of the regression of the
        total GNS against the GNS of each layer type, respectively, as
        a function of the same \ac{ema} alpha values. The total GNS (black) on
        the left is predicted well by individual layer types as indicated by the
        correlation coefficients (right), however the type with slope closest to
        1 is LayerNorm (center), only overestimating the GNS by less than 40\%
        across EMA alpha values.}
    \label{fig:regress_gns}
\end{figure}

So, when quantifying the relationship between the GNS of different layers, it
must be compared for different smoothing factors. Here, we show the regression
coefficients with respect to the alpha of the \ac{ema} filter in
Figure~\ref{fig:regress_gns}. The results show that the GNS of the LayerNorm
and Attention layers are highly predictive of the total GNS of the model.
In both cases, the slope is approximately 1.4, meaning the total GNS is
approximately 1.4 times the GNS of the LayerNorm or Attention layers.

Comparing the quality of this fit versus the quality of prior work's overall fit of the \ac{gns} to the critical batch size
(measured empirically)~\citep{mccandlish2018empirical}, the quality seems acceptable and
we do not need to apply this 1.4x correction factor, rather we just note that the true
$\Bcrit$ may be greater than the measured $\Bsimple$.

\section{Batch Size Scheduling}\label{batch-size}

We focus on two concerns that affect the practicality of batch size scheduling.
First, measuring the appropriate batch size without incurring any additional
training time. We find this is possible with the method described in Section~\ref{universal}.
Second, whether batch size scheduling is effective in practice. We find
it can offer significant savings in the required number of tokens processed
in Section~\ref{bss}.

\subsection{Universal GNS with Zero Overhead}\label{universal}

Capturing a GNS estimate for a linear layer is powerful, but efficiently
doing so presents a challenge. Such an estimate requires accumulating
per-example gradients of $ \text{hidden\_size}^2 $ across the sequence
dimension, compared to just $ \text{hidden\_size} $ with LayerNorm. This
increased size requires using more complex reductions in the kernel, rather than
a simple warp reduction followed by shared-memory atomic reduction with a final
atomic global reduction (as we can implement for LayerNorm per-example gradients
within shared memory). In addition, linear layer kernels are already highly
optimized and require using advanced techniques to keep GPU tensor cores fed
with data, so combining such a kernel with per-example gradient computation
- with its own memory overheads and corresponding available bandwidth reduction
- would be a difficult undertaking.

\begin{figure}
    \centering
    \includegraphics[width=0.6\textwidth]{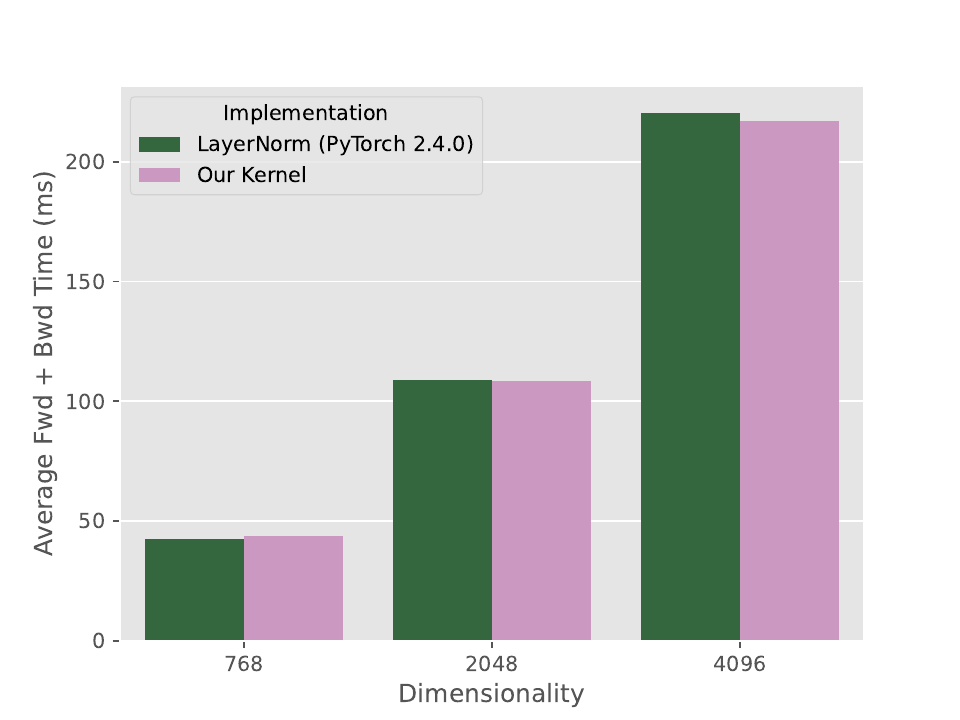}
    \caption{%
    Comparison of average time taken for a LayerNorm forward and backward pass
    with gradient accumulation when using PyTorch's native implementation versus
    our custom kernel computing per-example gradient norms in tandem. Measured
    on an Nvidia H100 GPU.}
    \label{fig:pt-vs-kernel}
\end{figure}

We thus implemented a LayerNorm-specific CUDA kernel that also captures GNS\@.
In experiments with language models at different scales, illustrated in
Figure~\ref{fig:pt-vs-kernel}, we find this kernel has practically zero
overhead compared to PyTorch's LayerNorm implementation. The complete source
code for this kernel is provided with the accompanying code for this
paper\footnote{\url{https://github.com/CerebrasResearch/nanoGNS/tree/main/exact/normgnorm}}.

\subsection{Case Study: Batch Size Schedule}\label{bss}

As a case study we continue with the 111M parameter language model on
OpenWebText described above. Over three seeds, we run both a fixed batch size
and a batch size schedule that increases linearly with the number of tokens
processed to the original batch size. We vary the batch size during training by
varying the number of gradient accumulation steps.

\begin{figure}
    \centering
    \includegraphics[width=\textwidth]{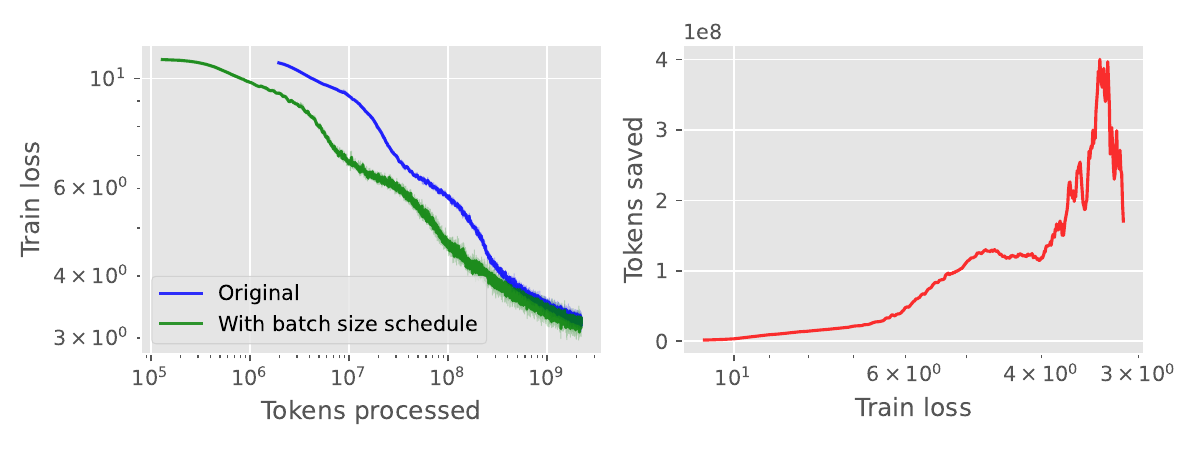}
    \caption{%
        (Left) Linear batch size schedule tracking the GNS over 2.2 billion
        tokens processed. Loss is plotted over a smoothed range from 3 runs
        using different Seeds. (Right) The number of tokens saved over the fixed
        batch size run to achieve the same loss.}
    \label{fig:111M_bss}
\end{figure}

The results of this experiment are shown in Figure~\ref{fig:111M_bss}. The left
plot shows the progression of the loss for both models, with the range of values
captured over different seeds. The mean loss for the linear batch size schedule
leads the fixed batch size throughout training. On the right, this lead is
quantified by interpolating the number of tokens saved to achieve the same loss.
The precise schedule used is shown in Figure~\ref{fig:111M_schedule} in
Appendix~\ref{app-bss}.

\section{Limitations}\label{limitations}

In this paper, we only studied Transformers, which include Normalization
sub-layers natively.  While Transformers are ubiquitous in machine
learning, there are many models, including variations of RNNs, CNNs,
and state-space models, that do not use such layers conventionally.
However, we note LayerNorm could be added to these networks with very
little overhead (in fact, the desire to normalize activations in RNNs
was one of the original motivations for developing LayerNorm;
application of batch normalization~\cite{ioffe2015batch} to RNNs was
``not obvious''~\cite{ba2016layer}).
Nevertheless, investigating LayerNorm-based \ac{gns} in these other
models requires further work.

Our work is also part of efforts to improve efficiency and address the
increasing costs of training and tuning large neural
networks~\cite{bender2021dangers}.
We provide both a more-efficient technique for computing the GNS, and
also, by enabling use of GNS statistics, we support compute-efficient
training recipes, such as use of dynamic batch sizes.
While some have argued that hyperscalers may re-invest any efficiency
savings into ever-larger models~\cite{patterson2021carbon}, for
academic researchers, such savings could allow pushing the
state-of-the-art, while still getting results in a reasonable
timeframe.
Recent efforts to enable frontier-model-performance within academic
budgets are encouraging, both to reduce
memory~\cite{malladi2023fine,dettmers2023qlora} and save compute
\cite{li2023clipa,anagnostidis2023navigating}.
Of course, even for such economical approaches, ``extensive
hyperparameter search'' may still be required~\cite{izsak2021train}.
There is a growing awareness that hyperparameter tuning has a negative
impact on equity in AI research, as tuning success depends directly on
researcher finances~\cite{strubell2019energy}.
A correlated trend is to use better training measurements (such as
gradient noise in batch and step size optimizers
(Section~\ref{related-work})) to reduce dependence on hyperparameters,
and in this way we hope our work can also ultimately improve research
equity.

\section{Conclusion}\label{conclusion}

This work set out to provide a practical method for computing the per-example
gradient norms necessary to compute the \ac{gns} independent of the training
configuration. In the process we discovered that not all the layers are
necessary for a practical estimate of the \ac{gns} and that the per-example
gradient norms can be computed for the normalization layers with zero overhead.
This enabled practical experiments, such as a batch size schedule and
replicating prior \ac{gns} observations. We are hopeful that democratising
access to \ac{gns} statistics, on any device, will enable subsequent
discoveries.

\bibliography{references.bib}


\begin{thebibliography}{63}


\ifx \showCODEN    \undefined \def \showCODEN     #1{\unskip}     \fi
\ifx \showDOI      \undefined \def \showDOI       #1{#1}\fi
\ifx \showISBNx    \undefined \def \showISBNx     #1{\unskip}     \fi
\ifx \showISBNxiii \undefined \def \showISBNxiii  #1{\unskip}     \fi
\ifx \showISSN     \undefined \def \showISSN      #1{\unskip}     \fi
\ifx \showLCCN     \undefined \def \showLCCN      #1{\unskip}     \fi
\ifx \shownote     \undefined \def \shownote      #1{#1}          \fi
\ifx \showarticletitle \undefined \def \showarticletitle #1{#1}   \fi
\ifx \showURL      \undefined \def \showURL       {\relax}        \fi
\providecommand\bibfield[2]{#2}
\providecommand\bibinfo[2]{#2}
\providecommand\natexlab[1]{#1}
\providecommand\showeprint[2][]{arXiv:#2}

\bibitem[Agarwal et~al\mbox{.}(2022)]%
        {agarwal2022estimating}
\bibfield{author}{\bibinfo{person}{Chirag Agarwal}, \bibinfo{person}{Daniel
  D'souza}, {and} \bibinfo{person}{Sara Hooker}.}
  \bibinfo{year}{2022}\natexlab{}.
\newblock \showarticletitle{Estimating example difficulty using variance of
  gradients}. In \bibinfo{booktitle}{\emph{Proceedings of the IEEE/CVF
  Conference on Computer Vision and Pattern Recognition}}.
  \bibinfo{pages}{10368--10378}.
\newblock


\bibitem[Anagnostidis et~al\mbox{.}(2023)]%
        {anagnostidis2023navigating}
\bibfield{author}{\bibinfo{person}{Sotiris Anagnostidis},
  \bibinfo{person}{Gregor Bachmann}, {and} \bibinfo{person}{Thomas Hofmann}.}
  \bibinfo{year}{2023}\natexlab{}.
\newblock \showarticletitle{Navigating Scaling Laws: Accelerating Vision
  Transformer's Training via Adaptive Strategies}.
\newblock \bibinfo{journal}{\emph{arXiv preprint arXiv:2311.03233}}
  (\bibinfo{year}{2023}).
\newblock


\bibitem[Ba et~al\mbox{.}(2022)]%
        {ba2022high}
\bibfield{author}{\bibinfo{person}{Jimmy Ba}, \bibinfo{person}{Murat~A.
  Erdogdu}, \bibinfo{person}{Taiji Suzuki}, \bibinfo{person}{Zhichao Wang},
  \bibinfo{person}{Denny Wu}, {and} \bibinfo{person}{Greg Yang}.}
  \bibinfo{year}{2022}\natexlab{}.
\newblock \bibinfo{title}{High-dimensional Asymptotics of Feature Learning: How
  One Gradient Step Improves the Representation}.
\newblock
\newblock
\showeprint[arxiv]{2205.01445}~[stat.ML]
\urldef\tempurl%
\url{https://arxiv.org/abs/2205.01445}
\showURL{%
\tempurl}


\bibitem[Ba et~al\mbox{.}(2016)]%
        {ba2016layer}
\bibfield{author}{\bibinfo{person}{Jimmy~Lei Ba}, \bibinfo{person}{Jamie~Ryan
  Kiros}, {and} \bibinfo{person}{Geoffrey~E. Hinton}.}
  \bibinfo{year}{2016}\natexlab{}.
\newblock \bibinfo{title}{Layer Normalization}.
\newblock
\newblock
\showeprint[arxiv]{1607.06450}~[stat.ML]


\bibitem[Balles et~al\mbox{.}(2016)]%
        {balles2016coupling}
\bibfield{author}{\bibinfo{person}{Lukas Balles}, \bibinfo{person}{Javier
  Romero}, {and} \bibinfo{person}{Philipp Hennig}.}
  \bibinfo{year}{2016}\natexlab{}.
\newblock \showarticletitle{Coupling adaptive batch sizes with learning rates}.
\newblock \bibinfo{journal}{\emph{arXiv preprint arXiv:1612.05086}}
  (\bibinfo{year}{2016}).
\newblock


\bibitem[Bender et~al\mbox{.}(2021)]%
        {bender2021dangers}
\bibfield{author}{\bibinfo{person}{Emily~M Bender}, \bibinfo{person}{Timnit
  Gebru}, \bibinfo{person}{Angelina McMillan-Major}, {and}
  \bibinfo{person}{Shmargaret Shmitchell}.} \bibinfo{year}{2021}\natexlab{}.
\newblock \showarticletitle{On the dangers of stochastic parrots: Can language
  models be too big?}. In \bibinfo{booktitle}{\emph{Proceedings of the 2021 ACM
  conference on fairness, accountability, and transparency}}.
  \bibinfo{pages}{610--623}.
\newblock


\bibitem[Black et~al\mbox{.}(2022)]%
        {black2022gptneox}
\bibfield{author}{\bibinfo{person}{Sidney Black}, \bibinfo{person}{Stella
  Biderman}, \bibinfo{person}{Eric Hallahan}, \bibinfo{person}{Quentin
  Anthony}, \bibinfo{person}{Leo Gao}, \bibinfo{person}{Laurence Golding},
  \bibinfo{person}{Horace He}, \bibinfo{person}{Connor Leahy},
  \bibinfo{person}{Kyle McDonell}, \bibinfo{person}{Jason Phang},
  \bibinfo{person}{Michael Pieler}, \bibinfo{person}{Usvsn~Sai Prashanth},
  \bibinfo{person}{Shivanshu Purohit}, \bibinfo{person}{Laria Reynolds},
  \bibinfo{person}{Jonathan Tow}, \bibinfo{person}{Ben Wang}, {and}
  \bibinfo{person}{Samuel Weinbach}.} \bibinfo{year}{2022}\natexlab{}.
\newblock \showarticletitle{{GPT-NeoX-20B}: An Open-Source Autoregressive
  Language Model}. In \bibinfo{booktitle}{\emph{Proceedings of the ACL Workshop
  on Challenges {\&} Perspectives in Creating Large Language Models}}.
\newblock


\bibitem[Bottou et~al\mbox{.}(2018)]%
        {bottou2018optimization}
\bibfield{author}{\bibinfo{person}{L{\'e}on Bottou}, \bibinfo{person}{Frank~E
  Curtis}, {and} \bibinfo{person}{Jorge Nocedal}.}
  \bibinfo{year}{2018}\natexlab{}.
\newblock \showarticletitle{Optimization methods for large-scale machine
  learning}.
\newblock \bibinfo{journal}{\emph{SIAM review}} \bibinfo{volume}{60},
  \bibinfo{number}{2} (\bibinfo{year}{2018}), \bibinfo{pages}{223--311}.
\newblock


\bibitem[Brown et~al\mbox{.}(2020)]%
        {brown2020language}
\bibfield{author}{\bibinfo{person}{Tom~B. Brown}, \bibinfo{person}{Benjamin
  Mann}, \bibinfo{person}{Nick Ryder}, \bibinfo{person}{Melanie Subbiah},
  \bibinfo{person}{Jared Kaplan}, \bibinfo{person}{Prafulla Dhariwal},
  \bibinfo{person}{Arvind Neelakantan}, \bibinfo{person}{Pranav Shyam},
  \bibinfo{person}{Girish Sastry}, \bibinfo{person}{Amanda Askell},
  \bibinfo{person}{Sandhini Agarwal}, \bibinfo{person}{Ariel Herbert-Voss},
  \bibinfo{person}{Gretchen Krueger}, \bibinfo{person}{Tom Henighan},
  \bibinfo{person}{Rewon Child}, \bibinfo{person}{Aditya Ramesh},
  \bibinfo{person}{Daniel~M. Ziegler}, \bibinfo{person}{Jeffrey Wu},
  \bibinfo{person}{Clemens Winter}, \bibinfo{person}{Christopher Hesse},
  \bibinfo{person}{Mark Chen}, \bibinfo{person}{Eric Sigler},
  \bibinfo{person}{Mateusz Litwin}, \bibinfo{person}{Scott Gray},
  \bibinfo{person}{Benjamin Chess}, \bibinfo{person}{Jack Clark},
  \bibinfo{person}{Christopher Berner}, \bibinfo{person}{Sam McCandlish},
  \bibinfo{person}{Alec Radford}, \bibinfo{person}{Ilya Sutskever}, {and}
  \bibinfo{person}{Dario Amodei}.} \bibinfo{year}{2020}\natexlab{}.
\newblock \bibinfo{title}{Language Models are Few-Shot Learners}.
\newblock
\newblock
\showeprint[arxiv]{2005.14165}~[cs.CL]
\urldef\tempurl%
\url{https://arxiv.org/abs/2005.14165}
\showURL{%
\tempurl}


\bibitem[Byrd et~al\mbox{.}(2012)]%
        {byrd2012sample}
\bibfield{author}{\bibinfo{person}{Richard~H Byrd}, \bibinfo{person}{Gillian~M
  Chin}, \bibinfo{person}{Jorge Nocedal}, {and} \bibinfo{person}{Yuchen Wu}.}
  \bibinfo{year}{2012}\natexlab{}.
\newblock \showarticletitle{Sample size selection in optimization methods for
  machine learning}.
\newblock \bibinfo{journal}{\emph{Mathematical programming}}
  \bibinfo{volume}{134}, \bibinfo{number}{1} (\bibinfo{year}{2012}),
  \bibinfo{pages}{127--155}.
\newblock


\bibitem[Chen et~al\mbox{.}(2018)]%
        {chen2018effect}
\bibfield{author}{\bibinfo{person}{Lingjiao Chen}, \bibinfo{person}{Hongyi
  Wang}, \bibinfo{person}{Jinman Zhao}, \bibinfo{person}{Dimitris
  Papailiopoulos}, {and} \bibinfo{person}{Paraschos Koutris}.}
  \bibinfo{year}{2018}\natexlab{}.
\newblock \showarticletitle{The effect of network width on the performance of
  large-batch training}.
\newblock \bibinfo{journal}{\emph{Advances in neural information processing
  systems}}  \bibinfo{volume}{31} (\bibinfo{year}{2018}).
\newblock


\bibitem[Choquet et~al\mbox{.}(1999)]%
        {choquet1999bootstrap}
\bibfield{author}{\bibinfo{person}{Denis Choquet}, \bibinfo{person}{Pierre
  L'Ecuyer}, {and} \bibinfo{person}{Christian L\'{e}ger}.}
  \bibinfo{year}{1999}\natexlab{}.
\newblock \showarticletitle{Bootstrap Confidence Intervals for Ratios of
  Expectations}.
\newblock \bibinfo{journal}{\emph{ACM Trans. Model. Comput. Simul.}}
  \bibinfo{volume}{9}, \bibinfo{number}{4} (\bibinfo{date}{oct}
  \bibinfo{year}{1999}), \bibinfo{pages}{326–348}.
\newblock
\showISSN{1049-3301}
\urldef\tempurl%
\url{https://doi.org/10.1145/352222.352224}
\showDOI{\tempurl}


\bibitem[Crowson(2024)]%
        {crowson2024kdiffusion}
\bibfield{author}{\bibinfo{person}{Katherine Crowson}.}
  \bibinfo{year}{2024}\natexlab{}.
\newblock \bibinfo{title}{k-diffusion}.
\newblock
  \bibinfo{howpublished}{\url{https://github.com/crowsonkb/k-diffusion}}.
\newblock
\newblock
\shownote{GitHub repository}.


\bibitem[Crowson et~al\mbox{.}(2024)]%
        {crowson2024scalable}
\bibfield{author}{\bibinfo{person}{Katherine Crowson},
  \bibinfo{person}{Stefan~Andreas Baumann}, \bibinfo{person}{Alex Birch},
  \bibinfo{person}{Tanishq~Mathew Abraham}, \bibinfo{person}{Daniel~Z Kaplan},
  {and} \bibinfo{person}{Enrico Shippole}.} \bibinfo{year}{2024}\natexlab{}.
\newblock \showarticletitle{Scalable High-Resolution Pixel-Space Image
  Synthesis with Hourglass Diffusion Transformers}.
\newblock \bibinfo{journal}{\emph{arXiv preprint arXiv:2401.11605}}
  (\bibinfo{year}{2024}).
\newblock


\bibitem[Dangel et~al\mbox{.}(2020)]%
        {dangel2020backpack}
\bibfield{author}{\bibinfo{person}{Felix Dangel}, \bibinfo{person}{Frederik
  Kunstner}, {and} \bibinfo{person}{Philipp Hennig}.}
  \bibinfo{year}{2020}\natexlab{}.
\newblock \showarticletitle{Back{PACK}: Packing more into Backprop}. In
  \bibinfo{booktitle}{\emph{International Conference on Learning
  Representations}}.
\newblock
\urldef\tempurl%
\url{https://openreview.net/forum?id=BJlrF24twB}
\showURL{%
\tempurl}


\bibitem[Dao et~al\mbox{.}(2022)]%
        {dao2022flash}
\bibfield{author}{\bibinfo{person}{Tri Dao}, \bibinfo{person}{Daniel~Y. Fu},
  \bibinfo{person}{Stefano Ermon}, \bibinfo{person}{Atri Rudra}, {and}
  \bibinfo{person}{Christopher Ré}.} \bibinfo{year}{2022}\natexlab{}.
\newblock \bibinfo{title}{FlashAttention: Fast and Memory-Efficient Exact
  Attention with IO-Awareness}.
\newblock
\newblock
\showeprint[arxiv]{2205.14135}~[cs.LG]
\urldef\tempurl%
\url{https://arxiv.org/abs/2205.14135}
\showURL{%
\tempurl}


\bibitem[De et~al\mbox{.}(2016)]%
        {de2016big}
\bibfield{author}{\bibinfo{person}{Soham De}, \bibinfo{person}{Abhay Yadav},
  \bibinfo{person}{David Jacobs}, {and} \bibinfo{person}{Tom Goldstein}.}
  \bibinfo{year}{2016}\natexlab{}.
\newblock \showarticletitle{Big batch {SGD}: Automated inference using adaptive
  batch sizes}.
\newblock \bibinfo{journal}{\emph{arXiv preprint arXiv:1610.05792}}
  (\bibinfo{year}{2016}).
\newblock


\bibitem[Dettmers et~al\mbox{.}(2023)]%
        {dettmers2023qlora}
\bibfield{author}{\bibinfo{person}{Tim Dettmers}, \bibinfo{person}{Artidoro
  Pagnoni}, \bibinfo{person}{Ari Holtzman}, {and} \bibinfo{person}{Luke
  Zettlemoyer}.} \bibinfo{year}{2023}\natexlab{}.
\newblock \showarticletitle{{QLoRA}: Efficient finetuning of quantized LLMs}.
\newblock \bibinfo{journal}{\emph{Advances in Neural Information Processing
  Systems}}  \bibinfo{volume}{36} (\bibinfo{year}{2023}).
\newblock


\bibitem[Dey et~al\mbox{.}(2023)]%
        {dey2023cerebrasgpt}
\bibfield{author}{\bibinfo{person}{Nolan Dey}, \bibinfo{person}{Gurpreet
  Gosal}, \bibinfo{person}{Zhiming}, \bibinfo{person}{Chen},
  \bibinfo{person}{Hemant Khachane}, \bibinfo{person}{William Marshall},
  \bibinfo{person}{Ribhu Pathria}, \bibinfo{person}{Marvin Tom}, {and}
  \bibinfo{person}{Joel Hestness}.} \bibinfo{year}{2023}\natexlab{}.
\newblock \bibinfo{title}{Cerebras-{GPT}: Open Compute-Optimal Language Models
  Trained on the {C}erebras Wafer-Scale Cluster}.
\newblock
\newblock
\showeprint[arxiv]{2304.03208}~[cs.LG]
\urldef\tempurl%
\url{https://arxiv.org/abs/2304.03208}
\showURL{%
\tempurl}


\bibitem[Duchi et~al\mbox{.}(2011)]%
        {duchi2011adaptive}
\bibfield{author}{\bibinfo{person}{John Duchi}, \bibinfo{person}{Elad Hazan},
  {and} \bibinfo{person}{Yoram Singer}.} \bibinfo{year}{2011}\natexlab{}.
\newblock \showarticletitle{Adaptive subgradient methods for online learning
  and stochastic optimization.}
\newblock \bibinfo{journal}{\emph{Journal of machine learning research}}
  \bibinfo{volume}{12}, \bibinfo{number}{7} (\bibinfo{year}{2011}).
\newblock


\bibitem[EleutherAI(2024)]%
        {eleuther2024gptneox}
\bibfield{author}{\bibinfo{person}{EleutherAI}.}
  \bibinfo{year}{2024}\natexlab{}.
\newblock \bibinfo{title}{GPT-NeoX}.
\newblock \bibinfo{howpublished}{\url{https://github.com/EleutherAI/gpt-neox}}.
\newblock
\newblock
\shownote{GitHub repository}.


\bibitem[Faghri et~al\mbox{.}(2020)]%
        {faghri2020study}
\bibfield{author}{\bibinfo{person}{Fartash Faghri}, \bibinfo{person}{David
  Duvenaud}, \bibinfo{person}{David~J Fleet}, {and} \bibinfo{person}{Jimmy
  Ba}.} \bibinfo{year}{2020}\natexlab{}.
\newblock \showarticletitle{A study of gradient variance in deep learning}.
\newblock \bibinfo{journal}{\emph{arXiv preprint arXiv:2007.04532}}
  (\bibinfo{year}{2020}).
\newblock


\bibitem[Gao et~al\mbox{.}(2020)]%
        {gao2020pile}
\bibfield{author}{\bibinfo{person}{Leo Gao}, \bibinfo{person}{Stella Biderman},
  \bibinfo{person}{Sid Black}, \bibinfo{person}{Laurence Golding},
  \bibinfo{person}{Travis Hoppe}, \bibinfo{person}{Charles Foster},
  \bibinfo{person}{Jason Phang}, \bibinfo{person}{Horace He},
  \bibinfo{person}{Anish Thite}, \bibinfo{person}{Noa Nabeshima},
  \bibinfo{person}{Shawn Presser}, {and} \bibinfo{person}{Connor Leahy}.}
  \bibinfo{year}{2020}\natexlab{}.
\newblock \bibinfo{title}{The Pile: An 800GB Dataset of Diverse Text for
  Language Modeling}.
\newblock
\newblock
\showeprint[arxiv]{2101.00027}~[cs.CL]


\bibitem[Gokaslan and Cohen(2019)]%
        {gokaslan2019owt}
\bibfield{author}{\bibinfo{person}{Aaron Gokaslan} {and} \bibinfo{person}{Vanya
  Cohen}.} \bibinfo{year}{2019}\natexlab{}.
\newblock \bibinfo{title}{OpenWebText Corpus}.
\newblock
  \bibinfo{howpublished}{\url{http://Skylion007.github.io/OpenWebTextCorpus}}.
\newblock


\bibitem[Golden et~al\mbox{.}(2024)]%
        {golden2024flash}
\bibfield{author}{\bibinfo{person}{Alicia Golden}, \bibinfo{person}{Samuel
  Hsia}, \bibinfo{person}{Fei Sun}, \bibinfo{person}{Bilge Acun},
  \bibinfo{person}{Basil Hosmer}, \bibinfo{person}{Yejin Lee},
  \bibinfo{person}{Zachary DeVito}, \bibinfo{person}{Jeff Johnson},
  \bibinfo{person}{Gu-Yeon Wei}, \bibinfo{person}{David Brooks}, {and}
  \bibinfo{person}{Carole-Jean Wu}.} \bibinfo{year}{2024}\natexlab{}.
\newblock \bibinfo{title}{Is Flash Attention Stable?}
\newblock
\newblock
\showeprint[arxiv]{2405.02803}~[cs.LG]


\bibitem[Goodfellow(2015)]%
        {goodfellow2015efficient}
\bibfield{author}{\bibinfo{person}{Ian Goodfellow}.}
  \bibinfo{year}{2015}\natexlab{}.
\newblock \bibinfo{title}{Efficient Per-Example Gradient Computations}.
\newblock
\newblock
\showeprint[arxiv]{1510.01799}~[stat.ML]
\urldef\tempurl%
\url{https://arxiv.org/abs/1510.01799}
\showURL{%
\tempurl}


\bibitem[Gray et~al\mbox{.}(2023)]%
        {gray2023efficient}
\bibfield{author}{\bibinfo{person}{Gavia Gray}, \bibinfo{person}{Anshul Samar},
  {and} \bibinfo{person}{Joel Hestness}.} \bibinfo{year}{2023}\natexlab{}.
\newblock \showarticletitle{Efficient and Approximate Per-Example Gradient
  Norms for Gradient Noise Scale}. In \bibinfo{booktitle}{\emph{Workshop on
  Advancing Neural Network Training: Computational Efficiency, Scalability, and
  Resource Optimization (WANT@NeurIPS 2023)}}.
\newblock
\urldef\tempurl%
\url{https://openreview.net/forum?id=xINTMAvPQA}
\showURL{%
\tempurl}


\bibitem[Hilton et~al\mbox{.}(2022)]%
        {hilton2022batch}
\bibfield{author}{\bibinfo{person}{Jacob Hilton}, \bibinfo{person}{Karl Cobbe},
  {and} \bibinfo{person}{John Schulman}.} \bibinfo{year}{2022}\natexlab{}.
\newblock \bibinfo{title}{Batch size-invariance for policy optimization}.
\newblock
\newblock
\showeprint[arxiv]{2110.00641}~[cs.LG]


\bibitem[Hoffmann et~al\mbox{.}(2022)]%
        {hoffmann2022training}
\bibfield{author}{\bibinfo{person}{Jordan Hoffmann}, \bibinfo{person}{Sebastian
  Borgeaud}, \bibinfo{person}{Arthur Mensch}, \bibinfo{person}{Elena
  Buchatskaya}, \bibinfo{person}{Trevor Cai}, \bibinfo{person}{Eliza
  Rutherford}, \bibinfo{person}{Diego de Las~Casas}, \bibinfo{person}{Lisa~Anne
  Hendricks}, \bibinfo{person}{Johannes Welbl}, \bibinfo{person}{Aidan Clark},
  \bibinfo{person}{Tom Hennigan}, \bibinfo{person}{Eric Noland},
  \bibinfo{person}{Katie Millican}, \bibinfo{person}{George van~den Driessche},
  \bibinfo{person}{Bogdan Damoc}, \bibinfo{person}{Aurelia Guy},
  \bibinfo{person}{Simon Osindero}, \bibinfo{person}{Karen Simonyan},
  \bibinfo{person}{Erich Elsen}, \bibinfo{person}{Jack~W. Rae},
  \bibinfo{person}{Oriol Vinyals}, {and} \bibinfo{person}{Laurent Sifre}.}
  \bibinfo{year}{2022}\natexlab{}.
\newblock \bibinfo{title}{Training Compute-Optimal Large Language Models}.
\newblock
\newblock
\showeprint[arxiv]{2203.15556}~[cs.CL]


\bibitem[Ioffe and Szegedy(2015)]%
        {ioffe2015batch}
\bibfield{author}{\bibinfo{person}{Sergey Ioffe} {and}
  \bibinfo{person}{Christian Szegedy}.} \bibinfo{year}{2015}\natexlab{}.
\newblock \showarticletitle{Batch normalization: Accelerating deep network
  training by reducing internal covariate shift}. In
  \bibinfo{booktitle}{\emph{International conference on machine learning}}.
  pmlr, \bibinfo{pages}{448--456}.
\newblock


\bibitem[Izsak et~al\mbox{.}(2021)]%
        {izsak2021train}
\bibfield{author}{\bibinfo{person}{Peter Izsak}, \bibinfo{person}{Moshe
  Berchansky}, {and} \bibinfo{person}{Omer Levy}.}
  \bibinfo{year}{2021}\natexlab{}.
\newblock \showarticletitle{How to train BERT with an academic budget}.
\newblock \bibinfo{journal}{\emph{arXiv preprint arXiv:2104.07705}}
  (\bibinfo{year}{2021}).
\newblock


\bibitem[Karras et~al\mbox{.}(2024)]%
        {karras2024edm2}
\bibfield{author}{\bibinfo{person}{Tero Karras}, \bibinfo{person}{Miika
  Aittala}, \bibinfo{person}{Jaakko Lehtinen}, \bibinfo{person}{Janne
  Hellsten}, \bibinfo{person}{Timo Aila}, {and} \bibinfo{person}{Samuli
  Laine}.} \bibinfo{year}{2024}\natexlab{}.
\newblock \bibinfo{title}{Analyzing and Improving the Training Dynamics of
  Diffusion Models}.
\newblock
\newblock
\showeprint[arxiv]{2312.02696}~[cs.CV]
\urldef\tempurl%
\url{https://arxiv.org/abs/2312.02696}
\showURL{%
\tempurl}


\bibitem[Kingma and Ba(2014)]%
        {kingma2014adam}
\bibfield{author}{\bibinfo{person}{Diederik~P Kingma} {and}
  \bibinfo{person}{Jimmy Ba}.} \bibinfo{year}{2014}\natexlab{}.
\newblock \showarticletitle{Adam: A method for stochastic optimization}.
\newblock \bibinfo{journal}{\emph{arXiv preprint arXiv:1412.6980}}
  (\bibinfo{year}{2014}).
\newblock


\bibitem[Kunstner et~al\mbox{.}(2023)]%
        {kunstner2023noise}
\bibfield{author}{\bibinfo{person}{Frederik Kunstner}, \bibinfo{person}{Jacques
  Chen}, \bibinfo{person}{Jonathan~Wilder Lavington}, {and}
  \bibinfo{person}{Mark Schmidt}.} \bibinfo{year}{2023}\natexlab{}.
\newblock \showarticletitle{Noise is not the main factor behind the gap between
  {SGD} and {Adam} on transformers, but sign descent might be}.
\newblock \bibinfo{journal}{\emph{arXiv preprint arXiv:2304.13960}}
  (\bibinfo{year}{2023}).
\newblock


\bibitem[Kunstner et~al\mbox{.}(2024)]%
        {kunstner2024heavy}
\bibfield{author}{\bibinfo{person}{Frederik Kunstner}, \bibinfo{person}{Robin
  Yadav}, \bibinfo{person}{Alan Milligan}, \bibinfo{person}{Mark Schmidt},
  {and} \bibinfo{person}{Alberto Bietti}.} \bibinfo{year}{2024}\natexlab{}.
\newblock \showarticletitle{Heavy-tailed class imbalance and why {Adam}
  outperforms gradient descent on language models}.
\newblock \bibinfo{journal}{\emph{arXiv preprint arXiv:2402.19449}}
  (\bibinfo{year}{2024}).
\newblock


\bibitem[Li et~al\mbox{.}(2022)]%
        {li2022large}
\bibfield{author}{\bibinfo{person}{Xuechen Li}, \bibinfo{person}{Florian
  Tramèr}, \bibinfo{person}{Percy Liang}, {and} \bibinfo{person}{Tatsunori
  Hashimoto}.} \bibinfo{year}{2022}\natexlab{}.
\newblock \bibinfo{title}{Large Language Models Can Be Strong Differentially
  Private Learners}.
\newblock
\newblock
\showeprint[arxiv]{2110.05679}~[cs.LG]
\urldef\tempurl%
\url{https://arxiv.org/abs/2110.05679}
\showURL{%
\tempurl}


\bibitem[Li et~al\mbox{.}(2023)]%
        {li2023clipa}
\bibfield{author}{\bibinfo{person}{Xianhang Li}, \bibinfo{person}{Zeyu Wang},
  {and} \bibinfo{person}{Cihang Xie}.} \bibinfo{year}{2023}\natexlab{}.
\newblock \showarticletitle{CLIPA-v2: Scaling CLIP Training with 81.1\%
  Zero-shot ImageNet Accuracy within a \$10,000 Budget; An Extra \$4,000
  Unlocks 81.8\% Accuracy}.
\newblock \bibinfo{journal}{\emph{arXiv preprint arXiv:2306.15658}}
  (\bibinfo{year}{2023}).
\newblock


\bibitem[Malladi et~al\mbox{.}(2023)]%
        {malladi2023fine}
\bibfield{author}{\bibinfo{person}{Sadhika Malladi}, \bibinfo{person}{Tianyu
  Gao}, \bibinfo{person}{Eshaan Nichani}, \bibinfo{person}{Alex Damian},
  \bibinfo{person}{Jason~D Lee}, \bibinfo{person}{Danqi Chen}, {and}
  \bibinfo{person}{Sanjeev Arora}.} \bibinfo{year}{2023}\natexlab{}.
\newblock \showarticletitle{Fine-tuning language models with just forward
  passes}.
\newblock \bibinfo{journal}{\emph{Advances in Neural Information Processing
  Systems}}  \bibinfo{volume}{36} (\bibinfo{year}{2023}),
  \bibinfo{pages}{53038--53075}.
\newblock


\bibitem[McCandlish et~al\mbox{.}(2018)]%
        {mccandlish2018empirical}
\bibfield{author}{\bibinfo{person}{Sam McCandlish}, \bibinfo{person}{Jared
  Kaplan}, \bibinfo{person}{Dario Amodei}, {and} \bibinfo{person}{OpenAI~Dota
  Team}.} \bibinfo{year}{2018}\natexlab{}.
\newblock \bibinfo{title}{An Empirical Model of Large-Batch Training}.
\newblock
\newblock
\showeprint[arxiv]{1812.06162}~[cs.LG]
\urldef\tempurl%
\url{https://arxiv.org/abs/1812.06162}
\showURL{%
\tempurl}


\bibitem[Miyato et~al\mbox{.}(2018)]%
        {miyato2018spectralnorm}
\bibfield{author}{\bibinfo{person}{Takeru Miyato}, \bibinfo{person}{Toshiki
  Kataoka}, \bibinfo{person}{Masanori Koyama}, {and} \bibinfo{person}{Yuichi
  Yoshida}.} \bibinfo{year}{2018}\natexlab{}.
\newblock \bibinfo{title}{Spectral Normalization for Generative Adversarial
  Networks}.
\newblock
\newblock
\showeprint[arxiv]{1802.05957}~[cs.LG]
\urldef\tempurl%
\url{https://arxiv.org/abs/1802.05957}
\showURL{%
\tempurl}


\bibitem[Nguyen et~al\mbox{.}(2019)]%
        {nguyen2019first}
\bibfield{author}{\bibinfo{person}{Thanh~Huy Nguyen}, \bibinfo{person}{Umut
  Simsekli}, \bibinfo{person}{Mert Gurbuzbalaban}, {and}
  \bibinfo{person}{Ga{\"e}l Richard}.} \bibinfo{year}{2019}\natexlab{}.
\newblock \showarticletitle{First exit time analysis of stochastic gradient
  descent under heavy-tailed gradient noise}.
\newblock \bibinfo{journal}{\emph{Advances in neural information processing
  systems}}  \bibinfo{volume}{32} (\bibinfo{year}{2019}).
\newblock


\bibitem[Patterson et~al\mbox{.}(2021)]%
        {patterson2021carbon}
\bibfield{author}{\bibinfo{person}{David Patterson}, \bibinfo{person}{Joseph
  Gonzalez}, \bibinfo{person}{Quoc Le}, \bibinfo{person}{Chen Liang},
  \bibinfo{person}{Lluis-Miquel Munguia}, \bibinfo{person}{Daniel Rothchild},
  \bibinfo{person}{David So}, \bibinfo{person}{Maud Texier}, {and}
  \bibinfo{person}{Jeff Dean}.} \bibinfo{year}{2021}\natexlab{}.
\newblock \showarticletitle{Carbon emissions and large neural network
  training}.
\newblock \bibinfo{journal}{\emph{arXiv preprint arXiv:2104.10350}}
  (\bibinfo{year}{2021}).
\newblock


\bibitem[Radford et~al\mbox{.}(2019)]%
        {radford2019language}
\bibfield{author}{\bibinfo{person}{Alec Radford}, \bibinfo{person}{Jeffrey Wu},
  \bibinfo{person}{Rewon Child}, \bibinfo{person}{David Luan},
  \bibinfo{person}{Dario Amodei}, \bibinfo{person}{Ilya Sutskever},
  {et~al\mbox{.}}} \bibinfo{year}{2019}\natexlab{}.
\newblock \showarticletitle{Language models are unsupervised multitask
  learners}.
\newblock \bibinfo{journal}{\emph{OpenAI blog}} \bibinfo{volume}{1},
  \bibinfo{number}{8} (\bibinfo{year}{2019}), \bibinfo{pages}{9}.
\newblock


\bibitem[Reddi et~al\mbox{.}(2019)]%
        {reddi2019convergence}
\bibfield{author}{\bibinfo{person}{Sashank~J Reddi}, \bibinfo{person}{Satyen
  Kale}, {and} \bibinfo{person}{Sanjiv Kumar}.}
  \bibinfo{year}{2019}\natexlab{}.
\newblock \showarticletitle{On the convergence of {Adam} and beyond}.
\newblock \bibinfo{journal}{\emph{arXiv preprint arXiv:1904.09237}}
  (\bibinfo{year}{2019}).
\newblock


\bibitem[Rochette et~al\mbox{.}(2019)]%
        {rochette2019efficient}
\bibfield{author}{\bibinfo{person}{Gaspar Rochette}, \bibinfo{person}{Andre
  Manoel}, {and} \bibinfo{person}{Eric~W. Tramel}.}
  \bibinfo{year}{2019}\natexlab{}.
\newblock \bibinfo{title}{Efficient Per-Example Gradient Computations in
  Convolutional Neural Networks}.
\newblock
\newblock
\showeprint[arxiv]{1912.06015}~[cs.LG]
\urldef\tempurl%
\url{https://arxiv.org/abs/1912.06015}
\showURL{%
\tempurl}


\bibitem[Schaul et~al\mbox{.}(2013)]%
        {schaul2013no}
\bibfield{author}{\bibinfo{person}{Tom Schaul}, \bibinfo{person}{Sixin Zhang},
  {and} \bibinfo{person}{Yann LeCun}.} \bibinfo{year}{2013}\natexlab{}.
\newblock \showarticletitle{No more pesky learning rates}. In
  \bibinfo{booktitle}{\emph{International conference on machine learning}}.
  PMLR, \bibinfo{pages}{343--351}.
\newblock


\bibitem[Shallue et~al\mbox{.}(2019)]%
        {shallue2019measuring}
\bibfield{author}{\bibinfo{person}{Christopher~J Shallue},
  \bibinfo{person}{Jaehoon Lee}, \bibinfo{person}{Joseph Antognini},
  \bibinfo{person}{Jascha Sohl-Dickstein}, \bibinfo{person}{Roy Frostig}, {and}
  \bibinfo{person}{George~E Dahl}.} \bibinfo{year}{2019}\natexlab{}.
\newblock \showarticletitle{Measuring the effects of data parallelism on neural
  network training}.
\newblock \bibinfo{journal}{\emph{Journal of Machine Learning Research}}
  \bibinfo{volume}{20}, \bibinfo{number}{112} (\bibinfo{year}{2019}),
  \bibinfo{pages}{1--49}.
\newblock


\bibitem[Simsekli et~al\mbox{.}(2019)]%
        {simsekli2019tail}
\bibfield{author}{\bibinfo{person}{Umut Simsekli}, \bibinfo{person}{Levent
  Sagun}, {and} \bibinfo{person}{Mert Gurbuzbalaban}.}
  \bibinfo{year}{2019}\natexlab{}.
\newblock \showarticletitle{A tail-index analysis of stochastic gradient noise
  in deep neural networks}. In \bibinfo{booktitle}{\emph{International
  Conference on Machine Learning}}. PMLR, \bibinfo{pages}{5827--5837}.
\newblock


\bibitem[Smith and Gray(2018)]%
        {smith2018opt_einsum}
\bibfield{author}{\bibinfo{person}{Daniel G.~A. Smith} {and}
  \bibinfo{person}{Johnnie Gray}.} \bibinfo{year}{2018}\natexlab{}.
\newblock \showarticletitle{opt\_einsum - A Python package for optimizing
  contraction order for einsum-like expressions}.
\newblock \bibinfo{journal}{\emph{Journal of Open Source Software}}
  \bibinfo{volume}{3}, \bibinfo{number}{26} (\bibinfo{year}{2018}),
  \bibinfo{pages}{753}.
\newblock
\urldef\tempurl%
\url{https://doi.org/10.21105/joss.00753}
\showDOI{\tempurl}


\bibitem[Smith and Le(2017)]%
        {smith2017bayesian}
\bibfield{author}{\bibinfo{person}{Samuel~L Smith} {and}
  \bibinfo{person}{Quoc~V Le}.} \bibinfo{year}{2017}\natexlab{}.
\newblock \showarticletitle{A {Bayesian} perspective on generalization and
  stochastic gradient descent}.
\newblock \bibinfo{journal}{\emph{arXiv preprint arXiv:1710.06451}}
  (\bibinfo{year}{2017}).
\newblock


\bibitem[Strubell et~al\mbox{.}(2019)]%
        {strubell2019energy}
\bibfield{author}{\bibinfo{person}{Emma Strubell}, \bibinfo{person}{Ananya
  Ganesh}, {and} \bibinfo{person}{Andrew McCallum}.}
  \bibinfo{year}{2019}\natexlab{}.
\newblock \showarticletitle{Energy and policy considerations for deep learning
  in {NLP}}.
\newblock \bibinfo{journal}{\emph{arXiv preprint arXiv:1906.02243}}
  (\bibinfo{year}{2019}).
\newblock


\bibitem[Wang et~al\mbox{.}(2016)]%
        {wang2016dueling}
\bibfield{author}{\bibinfo{person}{Ziyu Wang}, \bibinfo{person}{Tom Schaul},
  \bibinfo{person}{Matteo Hessel}, \bibinfo{person}{Hado Hasselt},
  \bibinfo{person}{Marc Lanctot}, {and} \bibinfo{person}{Nando Freitas}.}
  \bibinfo{year}{2016}\natexlab{}.
\newblock \showarticletitle{Dueling network architectures for deep
  reinforcement learning}. In \bibinfo{booktitle}{\emph{International
  conference on machine learning}}. PMLR, \bibinfo{pages}{1995--2003}.
\newblock


\bibitem[Wortsman et~al\mbox{.}(2023)]%
        {wortsman2023small}
\bibfield{author}{\bibinfo{person}{Mitchell Wortsman},
  \bibinfo{person}{Peter~J. Liu}, \bibinfo{person}{Lechao Xiao},
  \bibinfo{person}{Katie Everett}, \bibinfo{person}{Alex Alemi},
  \bibinfo{person}{Ben Adlam}, \bibinfo{person}{John~D. Co-Reyes},
  \bibinfo{person}{Izzeddin Gur}, \bibinfo{person}{Abhishek Kumar},
  \bibinfo{person}{Roman Novak}, \bibinfo{person}{Jeffrey Pennington},
  \bibinfo{person}{Jascha Sohl-dickstein}, \bibinfo{person}{Kelvin Xu},
  \bibinfo{person}{Jaehoon Lee}, \bibinfo{person}{Justin Gilmer}, {and}
  \bibinfo{person}{Simon Kornblith}.} \bibinfo{year}{2023}\natexlab{}.
\newblock \bibinfo{title}{Small-scale proxies for large-scale Transformer
  training instabilities}.
\newblock
\newblock
\showeprint[arxiv]{2309.14322}~[cs.LG]
\urldef\tempurl%
\url{https://arxiv.org/abs/2309.14322}
\showURL{%
\tempurl}


\bibitem[Yang et~al\mbox{.}(2021)]%
        {yang2022mup}
\bibfield{author}{\bibinfo{person}{Greg Yang}, \bibinfo{person}{Edward Hu},
  \bibinfo{person}{Igor Babuschkin}, \bibinfo{person}{Szymon Sidor},
  \bibinfo{person}{Xiaodong Liu}, \bibinfo{person}{David Farhi},
  \bibinfo{person}{Nick Ryder}, \bibinfo{person}{Jakub Pachocki},
  \bibinfo{person}{Weizhu Chen}, {and} \bibinfo{person}{Jianfeng Gao}.}
  \bibinfo{year}{2021}\natexlab{}.
\newblock \showarticletitle{{Tuning Large Neural Networks via Zero-Shot
  Hyperparameter Transfer}}. In \bibinfo{booktitle}{\emph{Advances in Neural
  Information Processing Systems}}.
\newblock


\bibitem[Yin et~al\mbox{.}(2018)]%
        {yin2018gradient}
\bibfield{author}{\bibinfo{person}{Dong Yin}, \bibinfo{person}{Ashwin
  Pananjady}, \bibinfo{person}{Max Lam}, \bibinfo{person}{Dimitris
  Papailiopoulos}, \bibinfo{person}{Kannan Ramchandran}, {and}
  \bibinfo{person}{Peter Bartlett}.} \bibinfo{year}{2018}\natexlab{}.
\newblock \showarticletitle{Gradient diversity: a key ingredient for scalable
  distributed learning}. In \bibinfo{booktitle}{\emph{International Conference
  on Artificial Intelligence and Statistics}}. PMLR,
  \bibinfo{pages}{1998--2007}.
\newblock


\bibitem[Yousefpour et~al\mbox{.}(2022)]%
        {opacus}
\bibfield{author}{\bibinfo{person}{Ashkan Yousefpour}, \bibinfo{person}{Igor
  Shilov}, \bibinfo{person}{Alexandre Sablayrolles}, \bibinfo{person}{Davide
  Testuggine}, \bibinfo{person}{Karthik Prasad}, \bibinfo{person}{Mani Malek},
  \bibinfo{person}{John Nguyen}, \bibinfo{person}{Sayan Ghosh},
  \bibinfo{person}{Akash Bharadwaj}, \bibinfo{person}{Jessica Zhao},
  \bibinfo{person}{Graham Cormode}, {and} \bibinfo{person}{Ilya Mironov}.}
  \bibinfo{year}{2022}\natexlab{}.
\newblock \bibinfo{title}{Opacus: User-Friendly Differential Privacy Library in
  PyTorch}.
\newblock
\newblock
\showeprint[arxiv]{2109.12298}~[cs.LG]


\bibitem[Zeiler(2012)]%
        {zeiler2012adadelta}
\bibfield{author}{\bibinfo{person}{Matthew~D Zeiler}.}
  \bibinfo{year}{2012}\natexlab{}.
\newblock \showarticletitle{Adadelta: an adaptive learning rate method}.
\newblock \bibinfo{journal}{\emph{arXiv preprint arXiv:1212.5701}}
  (\bibinfo{year}{2012}).
\newblock


\bibitem[Zhai et~al\mbox{.}(2023)]%
        {zhai2023stabilizing}
\bibfield{author}{\bibinfo{person}{Shuangfei Zhai}, \bibinfo{person}{Tatiana
  Likhomanenko}, \bibinfo{person}{Etai Littwin}, \bibinfo{person}{Dan
  Busbridge}, \bibinfo{person}{Jason Ramapuram}, \bibinfo{person}{Yizhe Zhang},
  \bibinfo{person}{Jiatao Gu}, {and} \bibinfo{person}{Josh Susskind}.}
  \bibinfo{year}{2023}\natexlab{}.
\newblock \bibinfo{title}{Stabilizing Transformer Training by Preventing
  Attention Entropy Collapse}.
\newblock
\newblock
\showeprint[arxiv]{2303.06296}~[cs.LG]
\urldef\tempurl%
\url{https://arxiv.org/abs/2303.06296}
\showURL{%
\tempurl}


\bibitem[Zhang and Sennrich(2019)]%
        {zhang2019root}
\bibfield{author}{\bibinfo{person}{Biao Zhang} {and} \bibinfo{person}{Rico
  Sennrich}.} \bibinfo{year}{2019}\natexlab{}.
\newblock \bibinfo{title}{Root Mean Square Layer Normalization}.
\newblock
\newblock
\showeprint[arxiv]{1910.07467}~[cs.LG]


\bibitem[Zhang et~al\mbox{.}(2019)]%
        {zhang2019algorithmic}
\bibfield{author}{\bibinfo{person}{Guodong Zhang}, \bibinfo{person}{Lala Li},
  \bibinfo{person}{Zachary Nado}, \bibinfo{person}{James Martens},
  \bibinfo{person}{Sushant Sachdeva}, \bibinfo{person}{George~E. Dahl},
  \bibinfo{person}{Christopher~J. Shallue}, {and} \bibinfo{person}{Roger
  Grosse}.} \bibinfo{year}{2019}\natexlab{}.
\newblock \bibinfo{title}{Which Algorithmic Choices Matter at Which Batch
  Sizes? Insights From a Noisy Quadratic Model}.
\newblock
\newblock
\showeprint[arxiv]{1907.04164}~[cs.LG]


\bibitem[Zhang et~al\mbox{.}(2020)]%
        {zhang2020adaptive}
\bibfield{author}{\bibinfo{person}{Jingzhao Zhang},
  \bibinfo{person}{Sai~Praneeth Karimireddy}, \bibinfo{person}{Andreas Veit},
  \bibinfo{person}{Seungyeon Kim}, \bibinfo{person}{Sashank Reddi},
  \bibinfo{person}{Sanjiv Kumar}, {and} \bibinfo{person}{Suvrit Sra}.}
  \bibinfo{year}{2020}\natexlab{}.
\newblock \showarticletitle{Why are adaptive methods good for attention
  models?}
\newblock \bibinfo{journal}{\emph{Advances in Neural Information Processing
  Systems}}  \bibinfo{volume}{33} (\bibinfo{year}{2020}),
  \bibinfo{pages}{15383--15393}.
\newblock


\bibitem[Zhang et~al\mbox{.}(2022)]%
        {zhang2022adam}
\bibfield{author}{\bibinfo{person}{Yushun Zhang}, \bibinfo{person}{Congliang
  Chen}, \bibinfo{person}{Naichen Shi}, \bibinfo{person}{Ruoyu Sun}, {and}
  \bibinfo{person}{Zhi-Quan Luo}.} \bibinfo{year}{2022}\natexlab{}.
\newblock \showarticletitle{Adam can converge without any modification on
  update rules}.
\newblock \bibinfo{journal}{\emph{Advances in neural information processing
  systems}}  \bibinfo{volume}{35} (\bibinfo{year}{2022}),
  \bibinfo{pages}{28386--28399}.
\newblock


\bibitem[Zhu et~al\mbox{.}(2018)]%
        {zhu2018anisotropic}
\bibfield{author}{\bibinfo{person}{Zhanxing Zhu}, \bibinfo{person}{Jingfeng
  Wu}, \bibinfo{person}{Bing Yu}, \bibinfo{person}{Lei Wu}, {and}
  \bibinfo{person}{Jinwen Ma}.} \bibinfo{year}{2018}\natexlab{}.
\newblock \showarticletitle{The anisotropic noise in stochastic gradient
  descent: Its behavior of escaping from sharp minima and regularization
  effects}.
\newblock \bibinfo{journal}{\emph{arXiv preprint arXiv:1803.00195}}
  (\bibinfo{year}{2018}).
\newblock


\end{thebibliography}

\FloatBarrier
\clearpage
\appendix

\section{Taxonomy}\label{taxonomy}

\citet{gray2023efficient} included an prior version of this taxonomy in their
work.

The following taxonomy describes the different methods available to compute the
$\sqn{\Gsmall}$ necessary to compute the \ac{gns} as described in Section~\ref{gns}.
``Gradient norm cost'' below refers to the cost of computing the norm of the gradient
for all parameters in the model, which is typically orders of magnitude smaller
than the cost of forward or backward passes.

\begin{itemize}
    \item Microbatch: multiple $\sqn{\Gsmall}$ are computed over a set of microbatches
        \begin{itemize}
            \item DDP: Each $\sqn{\Gsmall}$ is computed before gradients are communicated between \ac{ddp} nodes~\citep{mccandlish2018empirical}.

                \textcolor{teal}{Pros:} Only gradient norm cost.

                \textcolor{orange}{Cons:} Variance tied to number of DDP nodes (see Figure~\ref{fig:variance}), can't be used on one node.
            \item Sequential: Each $\sqn{\Gsmall}$ are computed sequentially during gradient accumulation.

                \textcolor{teal}{Pros:} Only gradient norm cost.

                \textcolor{orange}{Cons:} Variance tied to the number of gradient accumulation steps.
        \end{itemize}
    \item Subbatch: During gradient accumulation, select $\sqn{\Gsmall}$ partway through.

        \textcolor{teal}{Pros:} Only gradient norm cost, easy to implement.

        \textcolor{orange}{Cons:} Higher variance than Microbatch as $\sqn{\Gsmall}$ is not averaged.
    \item Per-example:

        \textcolor{teal}{Pros:} Independent of gradient accumulation or DDP configuration, minimal variance.
        \begin{itemize}
            \item Exact:
                \begin{itemize}
                    \item $\sqn{\Gsmall}$ is computed directly by the per-example gradient trick~\citep{goodfellow2015efficient,li2022large}.

                    \textcolor{teal}{Pros:} Minimal cost in 2D regime.

                    \textcolor{orange}{Cons:} Redundant computation required in 3D regime.
                    \item $\sqn{\Gsmall}$ is computed in tandem with the parameter gradients using the method described in Section~\ref{sim-per-example}.

                    \textcolor{teal}{Pros:} No redundant computation.

                    \textcolor{orange}{Cons:} Expansion in memory causes slowdowns as described in Section~\ref{flops}.
                \end{itemize}
            \item Approximation: $\sqn{\Gsmall}$ is approximated by assuming input activations are normally distributed with mean zero~\citep{gray2023efficient}.

            \textcolor{teal}{Pros:} Fewer FLOPs than Exact methods.

            \textcolor{orange}{Cons:} Not exact.
        \end{itemize}
\end{itemize}

All of the methods described above can be measured either online or offline.
The description above focuses on the online case; i.e. measuring the gradient
norms during training. To use these methods offline: run the
models without performing weight updates and measure gradient norms the same way.
The estimators of Equation~\ref{eq:g-est} and~\ref{eq:s-est} can then be
aggregated using a mean rather than an EMA or by using a method to estimate
measurement uncertainty such as the jackknife mentioned in
Figure~\ref{fig:variance} (described in the context of \ac{gns} by
\citet[App.B]{gray2023efficient}). This can be useful to estimate how long to
run the offline estimate.

\section{Additional Simultaneous Per-Example Gradient Norm Computations}\label{app-sim}

Algorithms~\ref{alg:embedding-layer} and~\ref{alg:ln-layer} describe the process
for computing the per-example gradient norms for the embedding and LayerNorm layers,
which are typically the remaining layers in Transformer models. RMSNorm~\citep{zhang2019root}
is practically identical to LayerNorm in this case because the parameters the
gradient is computed wrt are in the affine transform, which is the same in
both layer types.

\begin{algorithm}
\caption{Layernorm Simultaneous Per-Example Gradient Norm Computation}
\label{alg:ln-layer}
\begin{algorithmic}[1]
\Require gradient tensor $\gB$ of shape $(B, ..., K)$, input activation tensor $\xB$ of shape $(B, ..., K)$
\Ensure gamma gradient tensor $\gammaB'$ of shape $(K,)$, mean of per-example squared norms $\sqn{\gammaB_b'}$, gradient tensor $\betaB'$ of shape $(K,)$, mean of per-example squared norms $\sqn{\betaB_b'}$
\State $\gammaB_b' \gets \text{einsum}(\mlq b...k,b...k \rightarrow bk \mrq, \xB, \gB)$
\State $\sB_{\gamma} \gets \text{einsum}(\mlq b k \rightarrow b \mrq, \gamma_b'^2)$
\State $\gammaB' \gets \text{einsum}(\mlq b k \rightarrow k \mrq, \gammaB_b')$
\State $\sqn{\gammaB_b'} \gets 1/B \times \text{einsum}(\sB_\gamma, \mlq b \rightarrow \mrq) \times B^2$ \# reduce by mean then apply correction
\State $\betaB_b' \gets \text{einsum}(\mlq b...k \rightarrow bk \mrq, \gB)$
\State $\sB_{\beta} \gets \text{einsum}(\mlq b k \rightarrow b \mrq, \betaB_b'^2)$
\State $\betaB' \gets \text{einsum}(\mlq b k \rightarrow k \mrq, \betaB_b')$
\State $\sqn{\betaB_b'} \gets 1/B \times \text{einsum}(\sB_\beta, \mlq b \rightarrow \mrq) \times B^2$ \# reduce by mean then apply correction
\State \Return $\gammaB', \sqn{\gammaB_b'}$, $\betaB', \sqn{\betaB_b'}$
\end{algorithmic}
\end{algorithm}

\begin{algorithm}
\caption{Embedding Layer Simultaneous Per-Example Gradient Norm Computation}
\label{alg:embedding-layer}
\begin{algorithmic}[1]
\Require gradient tensor $\gB$ of shape $(B, T, D)$, input id tensor $\xB$ of shape $(B, T)$, vocabulary size $V$
\Ensure weight gradient tensor $\wB'$ of shape $(V, D)$, mean of per-example squared norms $\sqn{\wB_b'}$
\State $\oB \gets \text{onehot}(\xB, V)$
\State $\wB_b' \gets \text{einsum}(\mlq b t v,b t d \rightarrow b v d \mrq, \oB, \gB)$
\State $\mathbf{s}_{w} \gets \text{einsum}(\mlq b v d \rightarrow b \mrq, \wB_b'^2)$
\State $\wB' \gets \text{einsum}(\mlq b v d \rightarrow v d \mrq, \wB_b')$
\State $\sqn{\wB_b'} \gets 1/B \times \text{einsum}(\mathbf{s}_w, \mlq b \rightarrow \mrq) \times B^2$ \# reduce by mean then apply correction
\State \Return $\wB', \sqn{\wB_b'}$
\end{algorithmic}
\end{algorithm}

\section{Language Model Experiment Details}\label{app-experiments}

As mentioned in the text, the code to run the experiments described in this paper
can be found at \url{https://github.com/CerebrasResearch/nanoGNS/tree/main/exact}.

\subsection{Optimality on OpenWebText}\label{optimal}

We chose to use the Cerebras-GPT~\citep{dey2023cerebrasgpt} recipes for
experiments as they are designed to be Chinchilla optimal. This means that each
model size should achieve the lowest possible loss for a given FLOP
budget~\citep{hoffmann2022training}. However, these recipes were tuned on the
Pile dataset~\citep{gao2020pile} and we used the OpenWebText dataset~\citep{gokaslan2019owt}
so that results could be replicated (Pile is no longer publicly available).

To verify that the training protocol is optimal on OpenWebText, we performed
a small study to illustrate how the performance would vary as we vary the size
and total tokens trained on. Model size was varied by changing the hidden size:
the 70M model has a hidden size of 576, the 111M model has a hidden size of 768
and the 161M model has a hidden size of 960. The token budget for each model
size was chosen to keep the total FLOPs constant.

\begin{figure}
    \centering
    \includegraphics[width=0.8\textwidth]{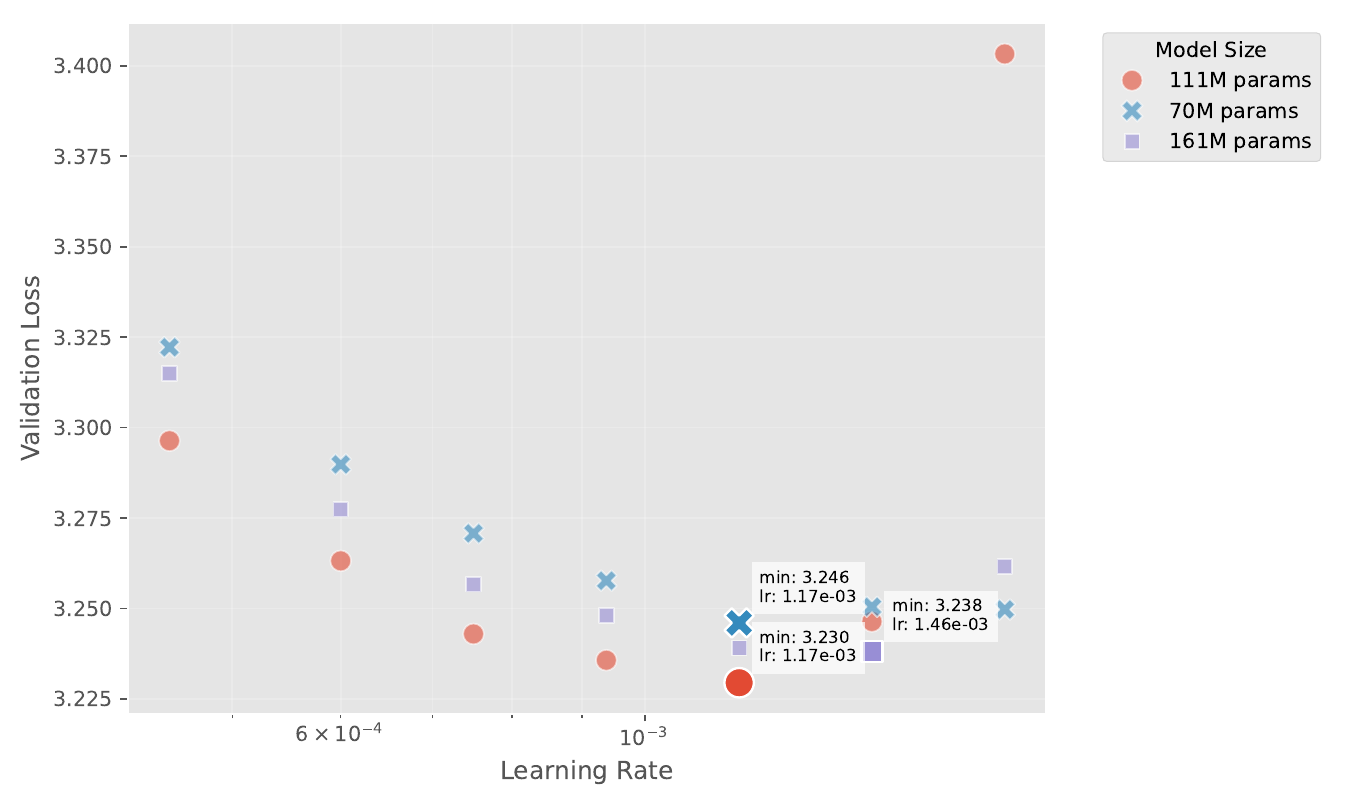}
    \caption{%
        The loss of models trained on OpenWebText with 70M, 111M and 161M
        parameters. The learning rate was varied to find the minima in the loss
        at each model scale. The optimal learning rate for each model size
        is annotated.}
    \label{fig:owt_optimal}
\end{figure}

The learning rate was varied to observe a minima in the loss at each model
scale. The results are shown in
Figure~\ref{fig:owt_optimal}. While we found that the learning rate may be
increased overall, the 111M model was found to have the lowest loss of the three
models. From these results we conclude that the training protocol is optimal
within this range of model sizes and we assume 111M is good enough. In other
words, a better model might exist between 70M and 161M parameters for this FLOP
budget but it isn't outside of this range.

\subsection{Flash Attention Numerical Instability}\label{instability}

The experiments described in Sections~\ref{experiments} and~\ref{bss} involve
Chinchilla optimal language models at a 111M scale~\citep{dey2023cerebrasgpt}.
These experiments were replicated according to the published information. We
encountered diverging runs when executing in bfloat16 \ac{amp} consistent with
the default settings in \href{https://github.com/karpathy/nanoGPT}{nanoGPT}.
These experiments were executed on NVIDIA A10 GPUs for accessible replication at
small scale. By ablation it was found that these runs would diverge:

\begin{itemize}
    \item Regardless of batch size schedule
    \item Regardless of hyperparameters: learning rate, weight decay, LayerNorm epsilon or Adam epsilon~\citep{wortsman2023small}
    \item When using PyTorch's AMP in bfloat16 precision
    \item When using Flash attention~\citep{dao2022flash,golden2024flash}
\end{itemize}

This was surprising because prior work had trained these models
successfully~\citep{dey2023cerebrasgpt}. In that work the model was also trained
using bfloat16 AMP precision, but it was trained on a Cerebras CS-2 system.
Due to this difference, we suspected the issue was due to a difference between
the efficient attention kernel and the Flash attention kernel in PyTorch.

\begin{figure}
    \centering
    \includegraphics[width=0.8\textwidth]{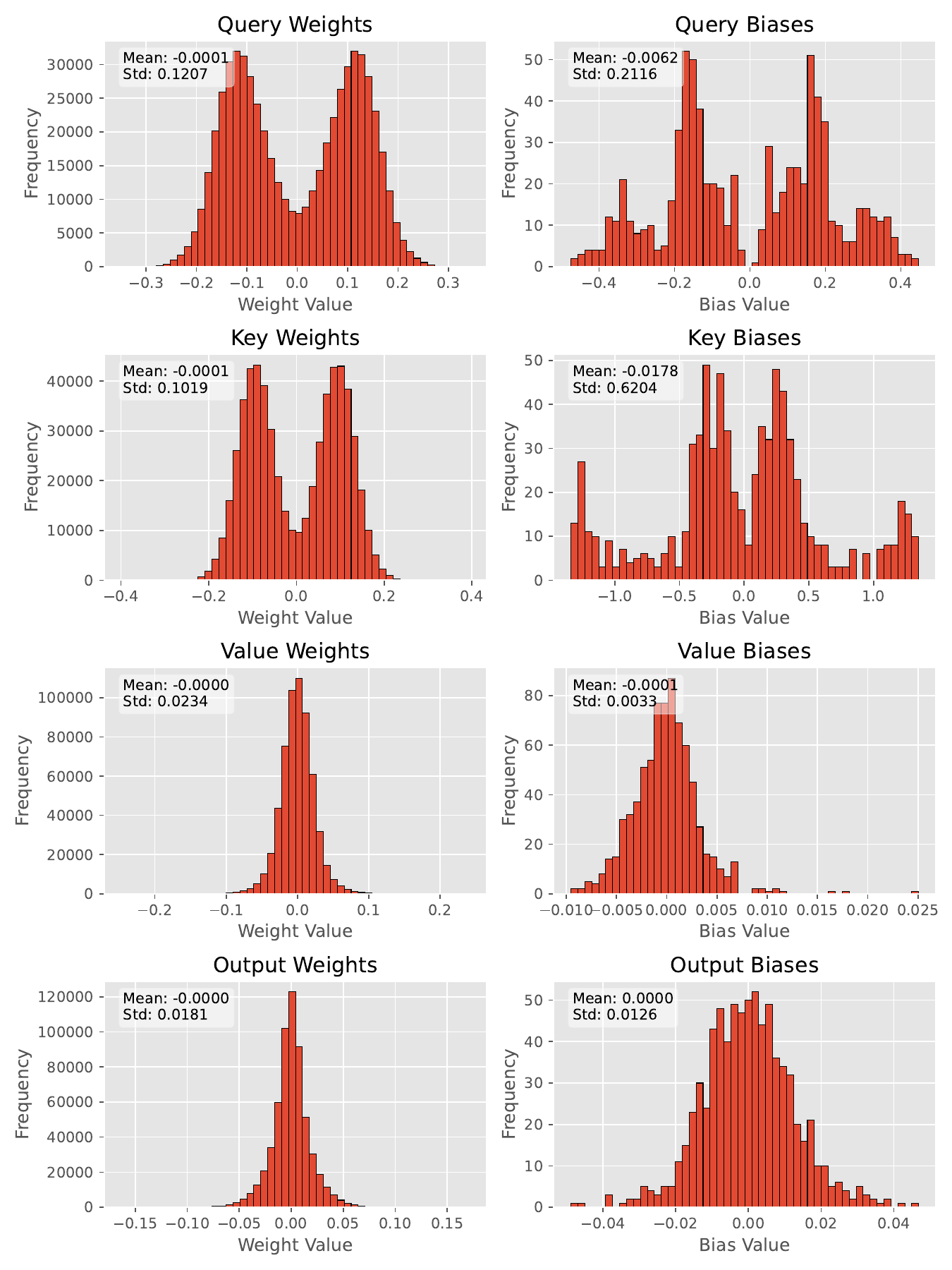}
    \caption{%
        Histograms of weights and biases for the 111M experiment described in
        Sections~\ref{experiments} and~\ref{bss} from the attention block
        containing the QKV projection, self-attention and output projection
        layers. The histograms for the query and key projection weights and
        biases are bimodal while the value projection weights and biases are
        not.}
    \label{fig:controls_optional}
\end{figure}

By inspecting the histograms of weights and biases in the Query, Key, Value
(QKV) projection during training, we found that range grew the fastest in block
1 (the \emph{second} block in the model). In addition, we
observed that the histogram of the query and key projection weights became
\emph{bimodal} as the gradient norm diverged. This is illustrated in
Figure~\ref{fig:controls_optional}. Further analysis of a checkpoint taken at
this point in training focused on the difference between gradients computed
using the flash attention kernel and the nanoGPT pure PyTorch attention
implementation using float32 precision. At initialization the gradients were not
significantly different but at the point of divergence there was a significant
difference coinciding with increased parameter norms in that layer.

\begin{figure}
    \centering
    \includegraphics[width=0.8\textwidth]{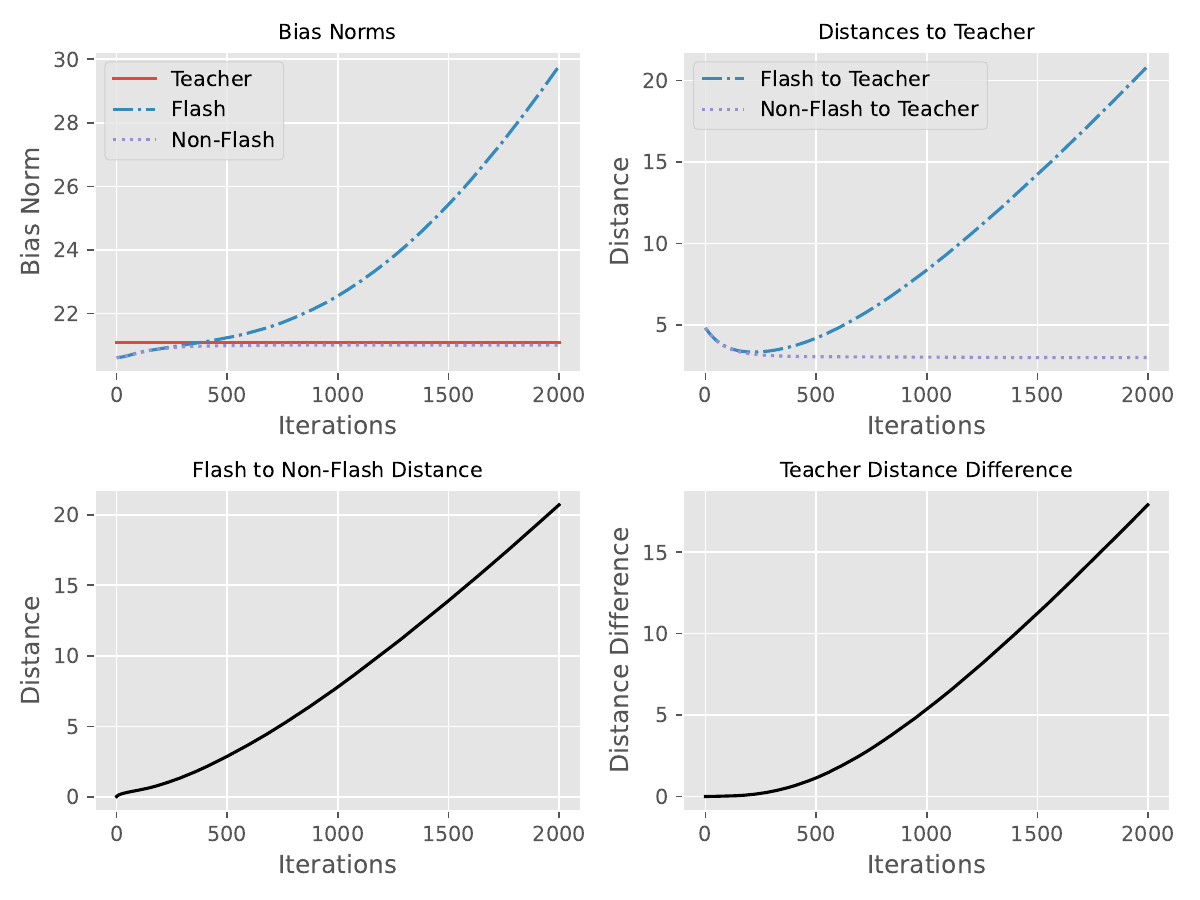}
    \caption{%
        Two ``student'' networks, identical to the ``teacher'' network except
        for the addition of a small amount of noise to the teacher's QKV
        projection bias. As training progresses, the student using Flash
        attention diverges for the same inputs. Plots are, clockwise from top
        left: ``Bias Norms'' shows the norms of the bias layer in each of the
        networks, ``Distances to Teacher'' shows the L2 distance from each
        student to the teacher. ``Flash to Non-Flash Distance'' shows the L2
        distance between the student using Flash attention and not,
        ``Teacher Distance Difference'' is the
        difference between the distances to the teacher for both cases.}
    \label{fig:flash_drift}
\end{figure}

To replicate the issue from scratch, we came up with a simulation from a generic
initialization. Inspired by the teacher-student experiment protocol proposed by
\citet{ba2022high} (although otherwise unrelated) we set up a learning task with
a ``teacher'' and ``student'' model with the same architecture. Both networks
begin with the same weights but we add a small amount of noise to the teacher's
weights. The student is trained to match the teacher's output. After
experimenting with hyperparameters we were able to replicated the divergence
seen during training\footnote{The code for this experiment is available
at
\url{https://gist.github.com/gaviag-cerebras/b77aef9de29e859a5e999a582d57f6a2}},
as illustrated in Figure~\ref{fig:flash_drift}.

\begin{figure}
    \centering
    \includegraphics[width=0.8\textwidth]{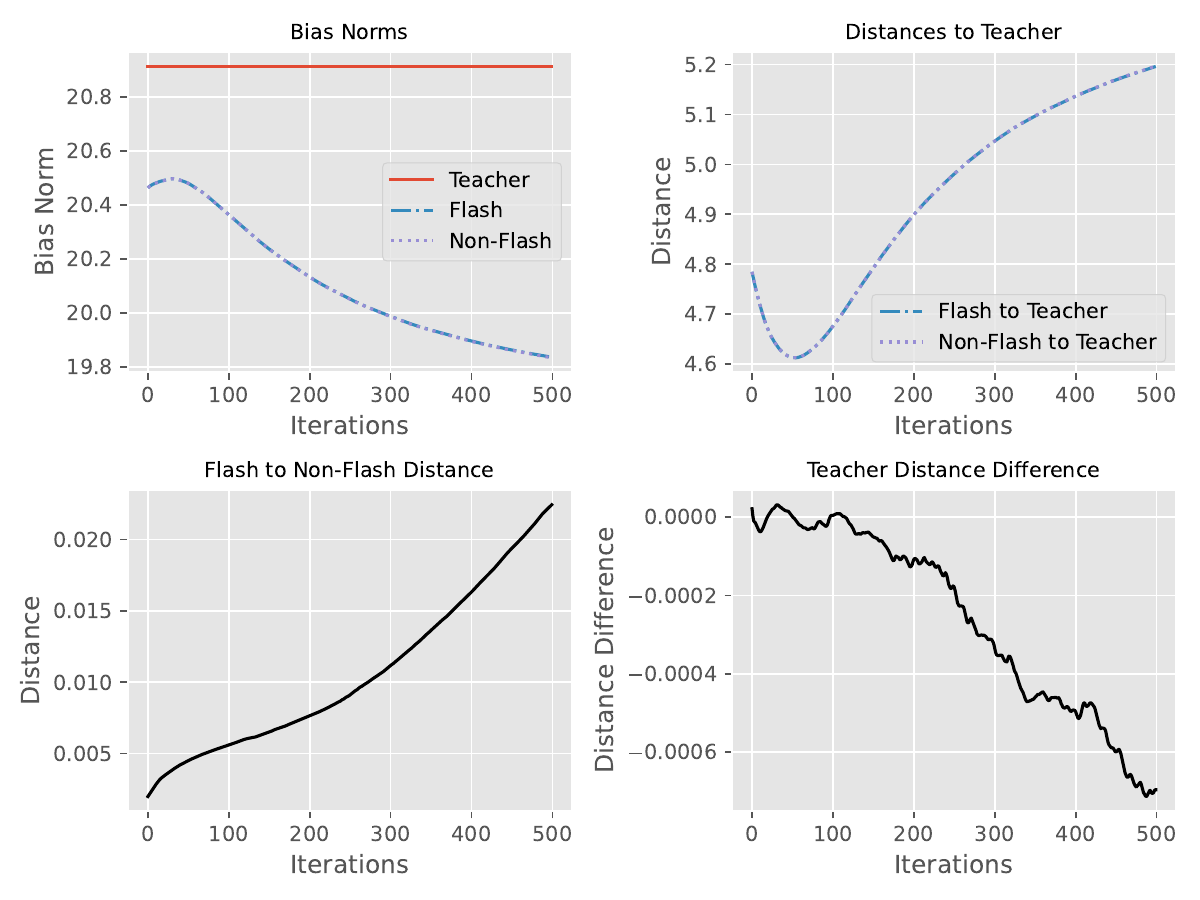}
    \caption{%
        Replication of the experiment described in Figure~\ref{fig:flash_drift}
        using cosine attention instead of Flash attention. The divergence observed
        no longer occurs.}
    \label{fig:flash_cosattn}
\end{figure}

Using this isolated simulation we were able to test different methods to
mitigate the divergence. \citet{karras2024edm2} suggested that cosine attention
could address similar divergences attributed to self-attention. In
Figure~\ref{fig:flash_cosattn} we replicated the experiment described in
Figure~\ref{fig:flash_drift} using cosine attention and found that the
divergence no longer occurred.

Separately, experimenting with precision ablation found that if float32
precision was used only in block 1 (2nd) then the divergence would also not
occur. Based on this and the above, we found the following two architectural
mitigations for the divergence, in \emph{only block 1 (2nd)}:

\begin{itemize}
    \item Use cosine attention, i.e. normalize the query and key head vectors before self-attention. \emph{OR}
    \item Use spectral normalization~\citep{miyato2018spectralnorm} on the QKV projection.
\end{itemize}

Critically, only modifying a single layer does not affect the throughput
of the model, the observed \ac{mfu} did not decrease by more than 1\% in either
case. Both of these bound the norm of the query and key head vectors prior to
attention. Spectral normalization achieves this because the QKV projection is
preceded by a LayerNorm layer. Using this mitigation on the 111M model allowed
the experiment to be replicated on an NVIDIA A10 GPU and we observed the same
behaviour as running the model more slowly in float32 precision.

Similar divergences are discussed in prior literature (and
\href{https://github.com/karpathy/nanoGPT/issues/137}{in nanoGPT's issue tracker})
but we are unable to verify that it is the same problem.
\citet{wortsman2023small} discuss how to build similar experiments to those
described above but do not investigate flash attention specifically.
\citet{golden2024flash} investigate the numerical stability of Flash attention
but neglect to demonstrate a failure mode that affects real training runs.
\citet{zhai2023stabilizing} focus on the numerical stability of attention in
general and propose a similar mitigation (their method, $\sigma$Reparam, is
a scaled version of spectral normalization) but do not investigate flash
attention specifically.

It is likely that the mitigation proposed will not work in all cases, such
as for larger models. However, we only needed to replicate at the scale we
were working at. The experiments in Figure~\ref{fig:flash_drift} and
Figure~\ref{fig:flash_cosattn} are included to illustrate how bounding the norm
of the query and key head vectors seems to be important for numerical stability.
However, this may change in future versions of the flash attention kernel, these
results were obtained with PyTorch 2.4.0.

\section{Additional GNS Results}\label{app-allgns}

\subsection{Additional GNS Phase Plot}\label{app-gns}

\begin{figure}
    \centering
    \includegraphics[width=\textwidth]{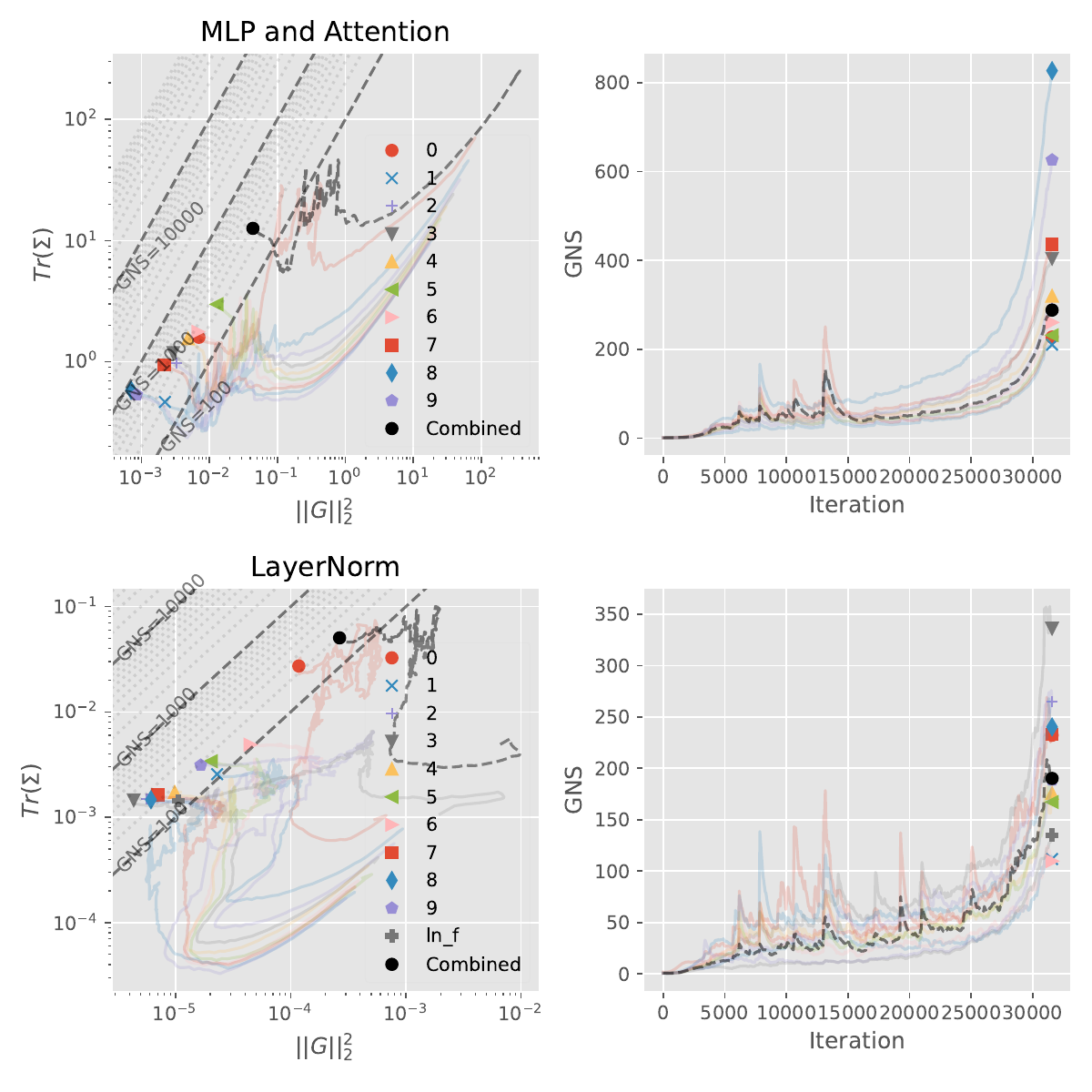}
    \caption{%
        GNS phase plot as in Figure~\ref{fig:gns_by_index} but focusing on the
        batch size schedule described in Section~\ref{bss}. Linear/Embedding
        layers are separated from LayerNorm layers by row and the component
        estimators of Equations~\ref{eq:g-est} and~\ref{eq:s-est} are plotted
        (left), with the GNS over the course of training (right).}
    \label{fig:gns_by_index_bss}
\end{figure}

Figure~\ref{fig:gns_by_index_bss} shows the GNS phase plot for the same model
as described in Section~\ref{experiments} but with the linear batch size schedule
described in Section~\ref{bss}.

\subsection{Batch Size Schedule}\label{app-bss}

\begin{figure}
     \centering
     \includegraphics[width=0.8\textwidth]{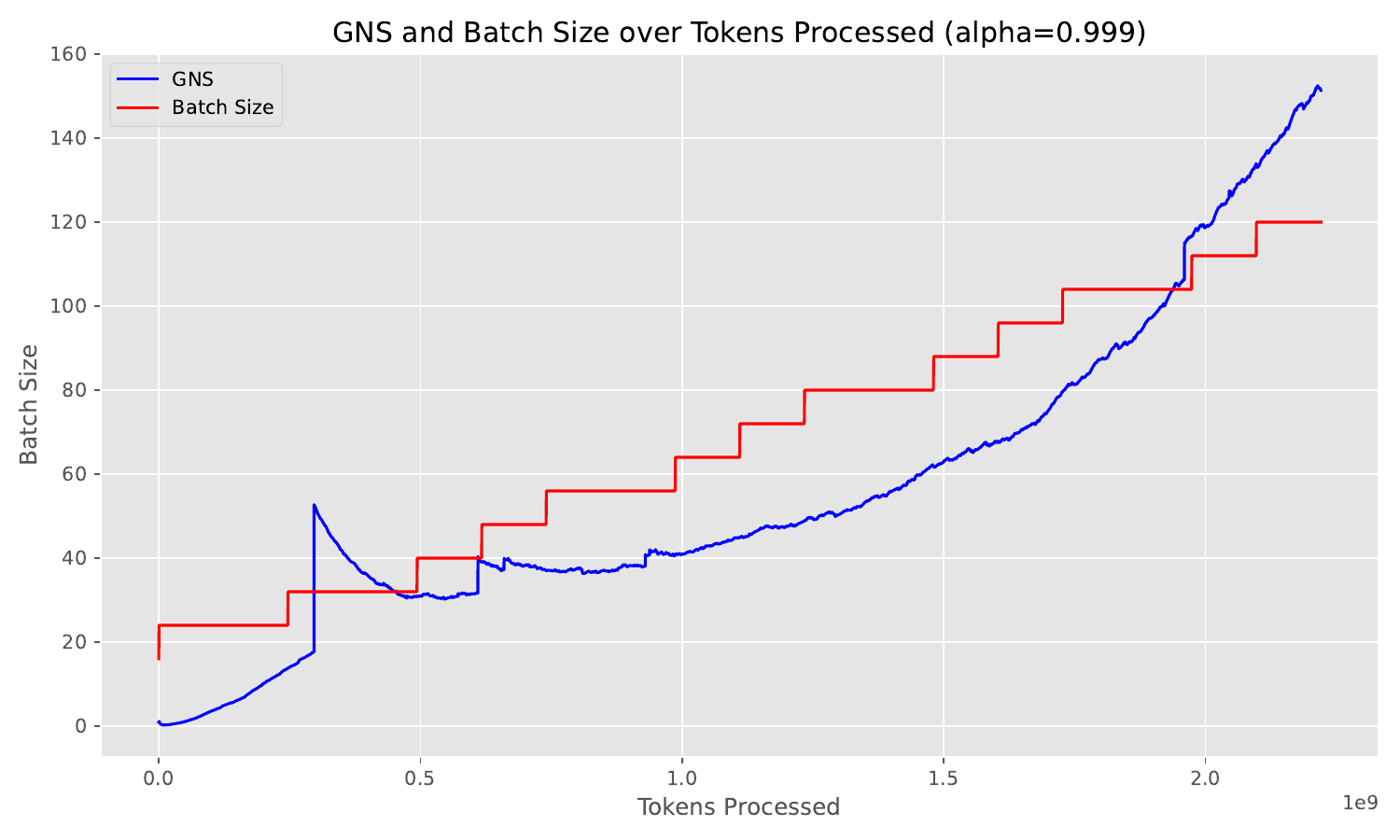}
     \caption{%
         The batch size schedule used and GNS observed in the 111M batch size
         schedule experiment illustrated in Figure~\ref{fig:111M_schedule}. An
         aliasing issue is noticeable in the interpolated linear batch size
         schedule that was used. This has since been fixed in the published
         code.}
     \label{fig:111M_schedule}
\end{figure}

The batch size schedule used in the experiment described in Section~\ref{bss} is
shown in Figure~\ref{fig:111M_schedule}.

\subsection{Larger Scale Training}\label{app-larger}

\begin{figure}
    \centering
    \begin{subfigure}[b]{0.49\textwidth}
        \centering
        \includegraphics[width=\linewidth]{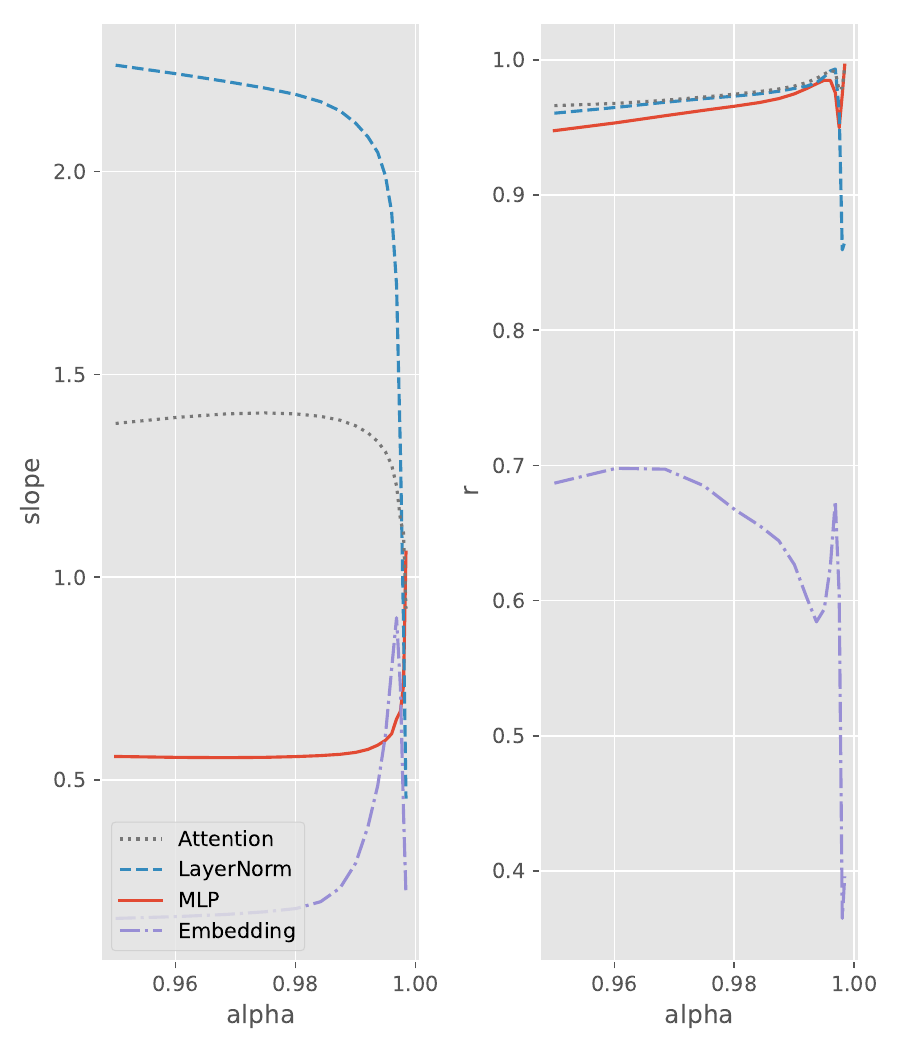}
        \caption{Regression analysis repeating Figure 6.}
        \label{fig:1p3b_left_subplot}
    \end{subfigure}
    \begin{subfigure}[b]{0.49\textwidth}
        \centering
        \includegraphics[width=\linewidth]{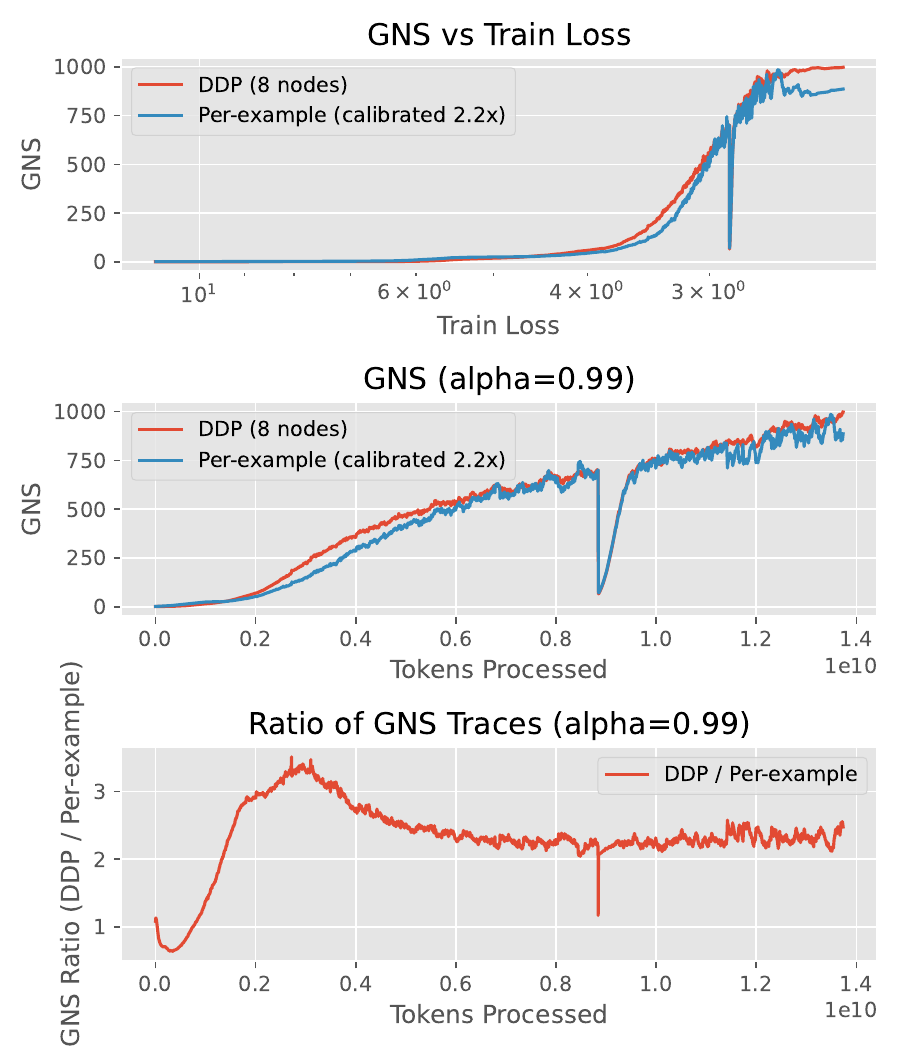}
        \caption{Comparison of GNS computed using traditional DDP methods and per-example gradient norms.}
        \label{fig:1p3b_right_subplot}
    \end{subfigure}
    \caption{%
          1.3B GPT model train on OpenWebText using 8 H100s, trained twice. (Left)
          Per-example gradient norms for all layers were gathered to replicate the analysis in Figure~\ref{fig:regress_gns}. (Right)
          Per-example gradient norms were gathered for only LayerNorm layers,
          then compared to the GNS computed using traditional DDP methods.}
    \label{fig:1p3b}
\end{figure}

To demonstrate that the method scales to larger models, we trained a 1.3B
parameter GPT
model\footnote{Again following the GPT2-like\citep{radford2019language} prescription from \citet{dey2023cerebrasgpt}.}
on OpenWebText using 8 H100 GPUs. The results of this
experiment are shown in Figure~\ref{fig:1p3b}. The left plot shows the
per-example gradient norms for all layers, while the right plot shows the
per-example gradient norms for only the LayerNorm layers. The GNS computed using
the traditional \ac{ddp} method is also shown for comparison. In
Figure~\ref{fig:1p3b_left_subplot} we observe that the LayerNorm remains
predictive of the total GNS, as in the 111M model results of
Figure~\ref{fig:regress_gns}. When all non-fused simultaneous per-example
gradient norms were collected we observed an \ac{mfu} of 40\% and when only the fused
LayerNorm layers were collected we observed an \ac{mfu} of 57\%.

After completing this experiment a bug was discovered in the code that decreased
per-example gradient norms by a constant factor. This caused an underestimation
of the \ac{gns}. In Figure~\ref{fig:1p3b_right_subplot} this can be seen when we
compare the \ac{gns} estimated via \ac{ddp} method. Initially, we assumed that
this constant factor was due a failure of the LayerNorm \ac{gns} approximation
to larger models. Unfortunately, we did not have the budget in time or resources
to rerun the experiment so we corrected the results by multiplying by the
constant factor observed in the comparison to the \ac{ddp} method.

This may be representative of real world scenarios where a large model is
pretrained over many \ac{ddp} nodes. As the user has access to two methods to
estimate the \ac{gns}, they may account for any bias or slope between the estimates.
Then, if it is necessary to continue training on a single node, they can use
the per-example gradient norms to estimate the \ac{gns}. Similar techniques can
involve enabling per-example \ac{gns} estimation for all layers for a short
time, or estimating the \ac{gns} offline as described in
Appendix~\ref{taxonomy}.

\section{FLOP \& I/O Formulae}\label{app-flops}

We use the following formulae in our FLOP and I/O cost estimations, where $B = \text{Batch Size}$, $T = \text{Sequence Length}$, $K = \text{Input Dimension}$, $L = \text{Output Dimension}$:
\begin{center}
\captionof{table}{FLOPs}
\begin{tabular}{ c || c | c }
\hline
Algorithm & Weight Gradient & Gradient Norms \\ \hline
Simultaneous & $B K L \left(2 T - 1\right) + K L \left(B - 1\right)$ & $B K L + B \left(K L - 1\right)$ \\
\cite{li2022large} & $K L \left(2 B T - 1\right)$ & $B T^{2} \cdot \left(2 K + 2 L - 2\right) + B T^{2}$ \\
\hline
\end{tabular}
\captionof{table}{I/O}
\begin{tabular}{ c || c | c }
\hline
Algorithm & Weight Gradient & Gradient Norms \\ \hline
Simultaneous & $B K L + B K T + B L T$ & $B K L + B$ \\
\cite{li2022large} & $B K T + B L T + K L$ & $2 B T^{2} + B$ \\
\hline
\end{tabular}
\end{center}

Solving the I/O equations above reproduces \cite{li2022large}'s analysis with $T = \frac{\sqrt{2} \sqrt{K L}}{2}$ at the cross-over point above which simultaneous calculation is more I/O efficient. Solving for FLOPs gives:
\[
T = \sqrt{\frac{2 K L - 1}{2 K + 2 L - 1}}.
\]

\FloatBarrier
\clearpage

\end{document}